\documentclass[journal,12pt,onecolumn,draftclsnofoot,]{IEEEtran}
\usepackage{times}

\usepackage{soul}
\usepackage{url}
\usepackage[utf8]{inputenc}
\usepackage{graphicx}
\usepackage{amsmath}
\usepackage{booktabs}
\usepackage{csquotes}
\usepackage{comment}
\usepackage{amsmath, bm}
\usepackage{amsfonts}
\usepackage{siunitx}
\usepackage{graphicx}
\usepackage{graphics}
\usepackage{layouts}
\usepackage{svg}
\usepackage{booktabs}
\usepackage{lipsum}
\usepackage{mathtools}
\usepackage{cuted}
\usepackage{changepage}
\usepackage{multirow}
\usepackage{xcolor}
\usepackage{subfig}

\usepackage[font=small,labelfont=bf]{caption}
\DeclareMathOperator*{\argmin}{arg\,min}
\graphicspath{ {./Figs/} }
\newcommand{\RNum}[1]{\uppercase\expandafter{\romannumeral #1\relax}}
\usepackage{tikz}
\newcommand*\circled[1]{\tikz[baseline=(char.base)]{\node[shape=circle,draw,inner sep=1.0pt] (char) {#1};}}
\newtheorem{definition}{Definition}

\begin{document}

\title{PRI-VAE: Principle-of-Relevant-Information Variational Autoencoders}

\author{Yanjun~Li,
        Shujian~Yu\IEEEauthorrefmark{1},
        Jose~C.~Principe,
        Xiaolin~Li,
        and~Dapeng~Wu\IEEEauthorrefmark{1}
\thanks{Yanjun~Li, Jose~C.~Principe and Dapeng~Wu are with the NSF Center for Big Learning, University of Florida, U.S.A (email: yanjun.li@ufl.edu, principe@cnel.ufl.edu and dpwu@ufl.edu)}
\thanks{Shujian~Yu is with the Machine Learning Group, NEC Laboratories Europe, Germany (email: Shujian.Yu@neclab.eu).}
\thanks{Xiaolin~Li is with the Cognization Lab (email: xiaolinli@ieee.org).}
\thanks{\IEEEauthorrefmark{1} To whom correspondence should be addressed.}
}

%
%

\markboth{}{}
%



\maketitle

\begin{abstract}
Although substantial efforts have been made to learn disentangled representations under the variational autoencoder (VAE) framework, the fundamental properties to the dynamics of learning of most VAE models still remain unknown and under-investigated. In this work, we first propose a novel learning objective, termed the principle-of-relevant-information variational autoencoder (PRI-VAE), to learn disentangled representations. We then present an information-theoretic perspective to analyze existing VAE models by inspecting the evolution of some critical information-theoretic quantities across training epochs. Our observations unveil some fundamental properties associated with VAEs. Empirical results also demonstrate the effectiveness of PRI-VAE on four benchmark data sets.
\end{abstract}


\begin{IEEEkeywords}
Disentangled representation learning, variational autoencoder, principle of relevant information.
\end{IEEEkeywords}

%
\IEEEpeerreviewmaketitle

\section{Introduction}
A central goal for representation learning models is that the resulting latent representation should be \emph{compact} yet \emph{disentangled}. \emph{Compact} requires the representation $\mathbf{z}$ does not contain any nuance factors in the input signal $\mathbf{x}$ that are not relevant for the desired response $\mathbf{y}$~\cite{achille2018information}, whereas \emph{disentangled} means that $\mathbf{z}$ is factorizable and has consistent semantics associated to different generating factors of the underlying data generation process. To achieve this goal in an unsupervised fashion, different variational autoencoder (VAE)~\cite{kingma2014auto} based models have been developed in recent years by formulating the objective as finding an informative latent representation that can minimize reconstruction error under proper regularizations. Well-known regularizations include the channel capacity~\cite{burgess2018understanding} and the total correlation~\cite{watanabe1960information} associated with each dimension of $\mathbf{z}$~\cite{kim2018disentangling,chen2018isolating}.


Despite these recent efforts, the fundamental properties of the dynamics of learning for most existing VAE models still remain unknown and under-investigated. Information theory provides a solid methodology to analyze the dynamics of learning and different trade-offs in deep neural networks (DNNs)~\cite{shwartz2017opening}. This is because the encoding, the decoding, and the compression are precisely among the core problems information theory was made to solve. Therefore, a natural idea is to analyze existing VAE models by inspecting the evolution of some key information-theoretic quantities associated with the latent representations across training epochs. Albeit easy-to-understand, estimating information-theoretic quantities (such as mutual information) exactly in high-dimensional space is problematic~\cite{paninski2003estimation}. As a result, current attempts (e.g.,~\cite{alemi2018information,achille2018information}) on interpreting VAE models still lie on using rate-distortion (RD) theory~\cite[Chapter~10]{cover2012elements} to approximately compare the RD trade-off of different VAEs or illuminating its connection to the famed Information Bottleneck (IB)~\cite{tishby99information}. 




In this work, we address aforementioned problems by deriving a novel VAE objective from a \emph{first principle}. We also suggest a novel estimator to evaluate mutual information and total correlation in VAEs; and illuminate its usage on analyzing and interpreting VAEs. Our contributions are fourfold:


\begin{itemize}
    \item We propose a novel VAE objective based on the principle of relevant information~\cite[Chapter~8]{principe2010information}. We term it PRI-VAE and establish its connections to other VAEs.
    \item We introduce the recently developed matrix-based R{\'e}nyi's $\alpha$-entropy functional estimator~\cite{giraldo2014measures,yu2019multivariate} to measure the mutual information $\mathbf{I}(\mathbf{x};\mathbf{z})$ and the total correlation $\mathbf{T}(\mathbf{z})$ of different VAEs across training epochs.
    \item We observed that, with the common stochastic gradient descent (SGD) optimization, both $\mathbf{I}(\mathbf{x};\mathbf{z})$ and $\mathbf{T}(\mathbf{z})$ increase rapidly at the first few epochs and then continuously decrease in the remaining epochs. We also observed a positive correlation between $\mathbf{I}(\mathbf{x};\mathbf{z})$ and $\mathbf{T}(\mathbf{z})$.
    \item Experiments on the four benchmark data sets suggest that PRI-VAE encourages more disentangled representations and reasonable reconstruction quality with fewer training epochs. We also demonstrate the flexibility of PRI-VAE and generalize its idea to a higher level.
\end{itemize}


\section{Preliminary Knowledge}
\subsection{VAE and the Evidence Lower Bound}

A VAE operates with two probabilistic mappings, an encoder $X\mapsto Z$ (represented by a neural network with parameter $\phi$), and a decoder ($Z\mapsto X$ represented by another neural network with parameter $\theta$). Normally, we assume a fixed prior distribution $p(\mathbf{z}
)$ over $\mathbf{z}$. Since the distribution of $\mathbf{x}$ is also fixed (i.e., the data distribution $q(\mathbf{x})$), the encoder and decoder induce joint distributions $q(\mathbf{x},\mathbf{z})=q_\phi(\mathbf{z}|\mathbf{x})q(\mathbf{x})$ and $p(\mathbf{x},\mathbf{z})=p_\theta(\mathbf{x}|\mathbf{z})p(\mathbf{z})$, respectively. An ideal VAE objective is to maximize the marginalized log-likelihood:
\begin{equation}\label{eq_ideal_VAE}
    \mathbb{E}_{p(\mathbf{x})}[\log p_\theta(\mathbf{x})].\\[3pt]
\end{equation}

Eq.~(\ref{eq_ideal_VAE}) is, however, not tractable and is approximated by the evidence lower bound (ELBO)~\cite{kingma2014auto}:
\begin{equation}\label{eq_VAE_obj}
    \mathbb{E}_{q(\mathbf{z}|\mathbf{x})}[\log p(\mathbf{x}|\mathbf{z})]-\mathbb{E}_{p(\mathbf{x})}[D_{KL}(q_\phi(\mathbf{z}|\mathbf{x})\|p(\mathbf{z}))], \\[3pt]
\end{equation}
where the first term measures the reconstruction loss, and the second one is the regularization term, which corresponds to the Kullback-Leibler (KL) divergence between the latent distribution $q_\phi(\mathbf{z}|\mathbf{x})$ and the prior distribution $p(\mathbf{z})$.
Usually, we assume $p(\mathbf{z})$ follows a standard isotropic multivariate normal distribution $\mathcal{N}(0,I)$ where $I$ is the identify matrix.


\subsection{PRI: the General Idea and its Objective}\label{sec:PRI_introduction}
PRI is an unsupervised information-theoretic principle that aims to perform mode decomposition of a random variable $X$ with a known (and fixed) probability distribution $g$. Suppose we obtain a reduced statistical representation characterized by a random variable $Y$ with probability distribution $f$. 

The PRI casts this problem as a trade-off between the entropy $\mathbf{H}(f)$ of $Y$ and its descriptive power about $X$ in terms of their divergence $D(f\|g)$:
\begin{equation}\label{eq_PRI_obj}
    J(f) = \argmin\limits_{f}\mathbf{H}(f) + \gamma D(f\|g),
\end{equation}
where $\gamma$ is a hyper-parameter controlling the amount of relevant information that $Y$ can extract from $X$. The minimization of entropy can be viewed as a means of reducing uncertainty (or redundancy) and finding the statistical regularities in the outcomes of a process, whereas the minimization of information divergence ensures that such regularities are closely related to $X$. The PRI is similar in spirit to the IB approach, but the formulation is different because PRI does not require a relevant auxiliary variable $Z$ and the optimization is done directly on the random variable $X$.

\begin{figure}
    \centering
    \subfloat[3d Gaussian.]{\includegraphics[width=0.28\textwidth]{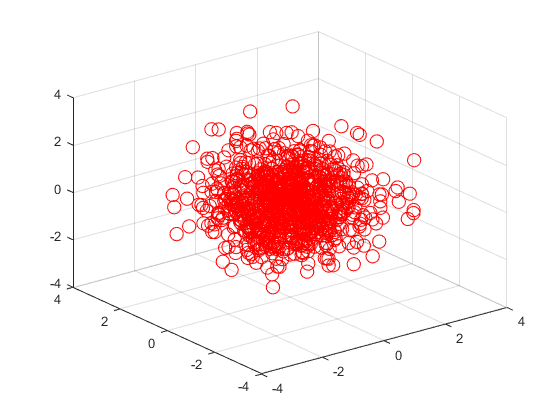}}
    \subfloat[$\gamma=0$]{\includegraphics[width=0.28\textwidth]{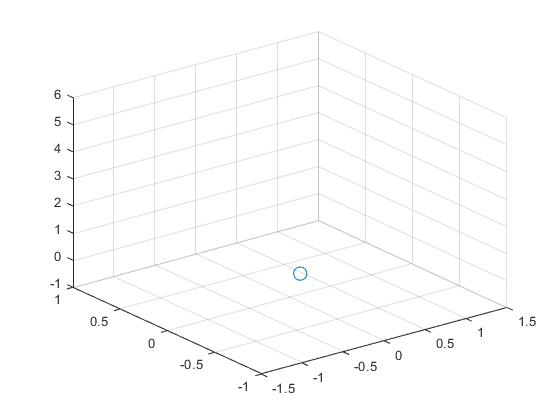}}
    \subfloat[$\gamma=1$]{\includegraphics[width=0.28\textwidth]{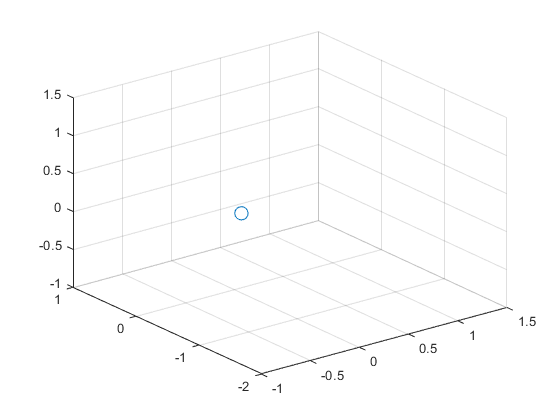}}\\
    \subfloat[$\gamma=2$]{\includegraphics[width=0.28\textwidth]{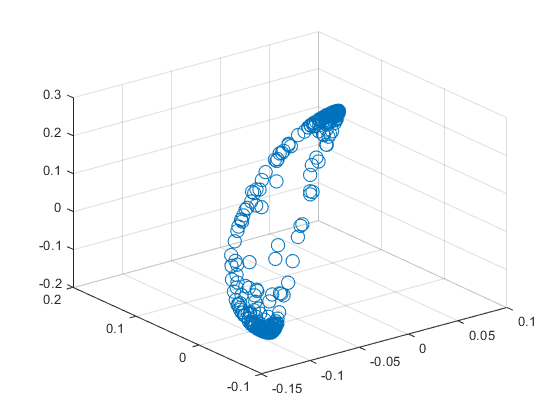}}
    \subfloat[$\gamma=5$]{\includegraphics[width=0.28\textwidth]{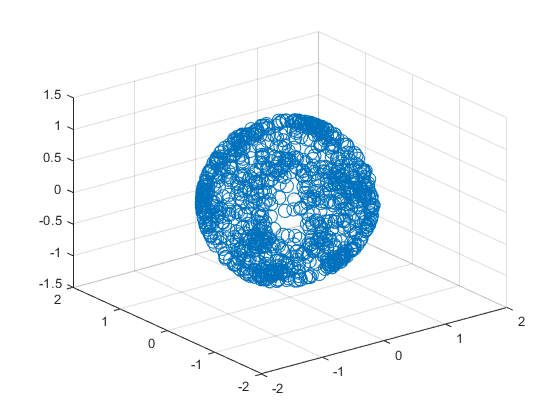}}
    \subfloat[$\gamma=100$]{\includegraphics[width=0.28\textwidth]{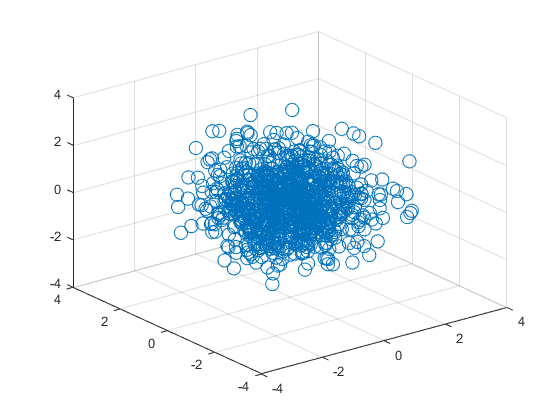}}
    \caption{Illustration of the structures revealed by the PRI for (a) a 3d isotropic Gaussian. As the values of $\gamma$ increase, the solution passes through (b) a single point, (c) mode, (d) principal curves, (e) principal surfaces, and in the extreme case of (f) $\gamma\rightarrow\infty$ we get back the data themselves as the solution.} 
    \label{fig:pri_demo}
\end{figure}

Note that the choice of entropy and divergence is application-specific and depends mostly on the simplicity of optimization. Although this work still uses the basic Shannon's differential entropy and the KL divergence for a fair comparison, the $2$-order R{\'e}nyi's entropy functional~\cite{renyi1961measures} and the Parzen window density estimator~\cite{parzen1962estimation} evidence a long track record of usefulness in various machine learning applications~\cite{principe2010information}.
\textcolor{black}{Fig.~\ref{fig:pri_demo} illustrates a set of solutions revealed by PRI that are related to the principal curves or surfaces.}

\section{PRI-VAE}
As mentioned earlier, a good latent representation $\mathbf{z}$ should be \emph{compact} yet \emph{disentangled}. We thus propose the following VAE objective:
\begin{equation}
\begin{split}
    \mathcal{L}_{\mathrm{PRI-VAE}} &= \mathbb{E}_{p(\mathbf{x})}\left[\mathbb{E}_{q(\mathbf{z} | \mathbf{x})}\left[\log p(\mathbf{x} | \mathbf{z})\right]\right. \\
    &- \alpha \mathbf{H}(\mathbf{z}) - \beta D_{KL}[q_{\phi}(\mathbf{z}) \| p(\mathbf{z})]. \\[3pt]
\end{split}
\label{eq:PRI_vae}
\end{equation}

Eq.~(\ref{eq:PRI_vae}) has three terms, but each term has its own physical meaning. 
$\mathbb{E}_{p(\mathbf{x})}\left[\mathbb{E}_{q(\mathbf{z} | \mathbf{x})}\left[\log p(\mathbf{x} | \mathbf{z})\right]\right.$ guarantees a reliable reconstruction of input $\mathbf{x}$ from latent representation $\mathbf{z}$.
$\alpha\mathbf{H}(\mathbf{z})$ regularizes the uncertainty or the degree of \emph{compactness} in $\mathbf{z}$: a large $\alpha$ encourages our model to learn a more compact latent representation, whereas a small $\alpha$ may lead to the latent representation contains nuisance factors that are irrelevant for downstream tasks. On the other hand, $\beta$ controls the closeness between $q_\phi(\mathbf{z})$ and $p(\mathbf{z})$, which can be interpreted as the extent of \emph{disentanglement} of $\mathbf{z}$: a large $\beta$ enforces the independence in each dimension of $\mathbf{z}$, and should, therefore, be preferred. One should note that, the combination of $\alpha \mathbf{H}(\mathbf{z})+\beta D_{KL}[q_\phi(\mathbf{z})\| p(\mathbf{z})]$ is exactly the objective of PRI, in which the ratio $\beta/\alpha$ plays the same role as $\gamma$ in Eq.~(\ref{eq_PRI_obj}). We therefore term Eq.~(\ref{eq:PRI_vae}) the principle-of-relevant-information variational autoencoder (PRI-VAE).


\subsection{Optimizing the PRI-VAE}
To optimize the PRI-VAE objective, we need to compute the inference marginal $q(\mathbf{z}) = \mathbb{E}_{p(\mathbf{x})}[q(\mathbf{z} | \mathbf{x})]$.
However, in practice, exactly computing its value is intractable in the training phase, since it depends on the entire data set. 
In PRI-VAE, we use the weighted Monte Carlo approximation suggested in~\cite{chen2018isolating} to estimate $q(\mathbf{z})$, as it does not require additional hyper-parameters or inner optimization loops. 



\subsection{Relation to Prior Art}\label{sec:PRI_VAE_relation}
In this section, we establish the connections between PRI-VAE and other recently developed VAEs.

$\beta$-VAE~\cite{higgins2017beta} learns disentangled representation by introducing a hyper-parameter $\beta$ (usually $\beta>1$) to heavily penalize the term
$D_{KL}(q_\phi(\mathbf{z}|\mathbf{x})\|p(\mathbf{z}))$ in Eq.~(\ref{eq_VAE_obj}). The emergence of disentanglement can be simply explained if we decompose $\mathbb{E}_{p(\mathbf{x})}[D_{KL}(q_\phi(\mathbf{z}|\mathbf{x})\|p(\mathbf{z}))]$ as (see proof in~\cite{kim2018disentangling}):
\begin{equation}\label{eq_decomp_KL}
    \underbrace{\mathbf{I}(\mathbf{x};\mathbf{z})}_{\circled{\tiny 1}}
    +\underbrace{D_{KL}(q_\phi(\mathbf{z})\|p(\mathbf{z})}_{\circled{\tiny 2}}). \\[3pt]
\end{equation}

To address this limitation, DIP-VAE~\cite{kumar2018variational} and InfoVAE~\cite{zhao2019infovae} suggest assigning different weights to the two terms in Eq.~(\ref{eq_decomp_KL}).
Moreover, suppose $p(\mathbf{z})$ is factorizable, \circled{\footnotesize 2} can be further decomposed as~\cite{kim2018disentangling,chen2018isolating}:
{\small
\begin{equation}\label{eq_tc_decompose}
\begin{split}
&\circled{\footnotesize 2} = 
\underbrace{D_{KL}(q(\mathbf{z})\|\prod\limits_{j=1}q(\mathbf{z}_j)}_{\circled{\tiny A}} + \sum_{d} \underbrace{\operatorname{D_{KL}}\left(q_{\phi}\left(\mathbf{z}_{j}\right) \| p\left(\mathbf{z}_{j}\right)\right)}_{\circled{\tiny B}}), \\[3pt]
\end{split}
\end{equation}
}
where term \circled{\footnotesize A} is the total correlation regularization associated with each dimension of $\mathbf{z}$ and term \circled{\footnotesize B} refers to the dimension-wise KL divergence ($d$ is the dimension of $\mathbf{z}$), which controls the penalty on the deviation between each latent dimension to the prior.
FactorVAE~\cite{kim2018disentangling} and $\beta$-TCVAE~\cite{chen2018isolating} add an external penalty on \circled{\footnotesize A} to encourage the independence in each dimension of $\mathbf{z}$.



Before our work, IB approach (e.g., \cite{achille2018information}) suggests a restriction on $\mathbf{I}(\mathbf{x};\mathbf{z})$ in Eq.~(\ref{eq_decomp_KL}) to keep $\mathbf{z}$ from containing nuance factors, whereas~\cite{alemi2018information,burgess2018understanding} introduce a hard constraint on the channel capacity to make sure $\mathbf{I}(\mathbf{x};\mathbf{z})$ is consistently larger than a predefined value.
PRI-VAE absorbs both proposals by flexibly adjusting the value of $\beta/\alpha$: a large $\alpha$ penalizes more on $\mathbf{H}(\mathbf{z})$ (and hence $\mathbf{I}(\mathbf{x};\mathbf{z})$, because $\mathbf{I}(\mathbf{x};\mathbf{z})$ is always saturated~\cite{esmaeili2019structured}) in practice, whereas a small $\alpha$ avoids $\mathbf{H}(\mathbf{z})$ reducing to zero, which also prevents a small channel capacity $\mathbf{I}(\mathbf{x};\mathbf{z})$. 
In an extreme case, when $\alpha=0$, we get back to the InfoVAE objective. 
We provide, in Table~\ref{tb:regularization}, an overview of objectives in different VAE models.

\begin{table}[h]
\caption{Objectives of different VAE models. $\protect\circled{\footnotesize 1}^{*}$ represents the $\mathbf{H}(\mathbf{z})$ term, which is a tight upper bound of the term \protect\circled{\footnotesize 1}.}
\label{tb:regularization}
\renewcommand{\arraystretch}{1.2}
\begin{center}
\scalebox{1.0}{
\begin{tabular}{lcl}
\hline
Method & \multicolumn{1}{l}{Reconstruction term} & Regularization term \\ \hline
$\beta$-VAE~\cite{higgins2017beta} & \multirow{7}{*}{$\mathbb{E}_{q(\mathbf{z}|\mathbf{x})}[\log p(\mathbf{x}|\mathbf{z})]$} & $\beta(\circled{\footnotesize 1} + \circled{\footnotesize 2})$ \\
AnnealedVAE~\cite{burgess2018understanding} &  & $\gamma |(\circled{\footnotesize 1} + \circled{\footnotesize 2}) - C|$ \\
DIP-VAE~\cite{kumar2018variational} &  & $\circled{\footnotesize 1} + \lambda \circled{\footnotesize 2}$ \\
InfoVAE~\cite{zhao2019infovae} &  & $\lambda\circled{\footnotesize 2}$ \\
FactorVAE~\cite{kim2018disentangling} &  & $\circled{\footnotesize 1} + \gamma\circled{\footnotesize A} + \circled{\footnotesize B}$ \\
$\beta$-TCVAE~\cite{chen2018isolating} &  & $\circled{\footnotesize 1} + \beta\circled{\footnotesize A} + \circled{\footnotesize B}$ \\ \cline{1-1} \cline{3-3} 
PRI-VAE &  & $\alpha \circled{\footnotesize 1}^{*} + \beta\circled{\footnotesize 2}$ \\ \hline
\end{tabular}}
\end{center}
\end{table}

\section{Empirical behavior of $\mathbf{I}(\mathbf{x};\mathbf{z})$ and $\mathbf{T}(\mathbf{z})$}\label{sec:Empirical_behavior}




Motivated by~\cite{shwartz2017opening}, we study disentanglement behavior and the learning dynamics of VAEs by measuring the evolution of $\mathbf{I}(\mathbf{x};\mathbf{z})$ and $\mathbf{T}(\mathbf{z})$ across training epochs. 

Given input $\mathbf{x}\in\mathbb{R}^p$ and latent representation $\mathbf{z}\in\mathbb{R}^d$, the standard Shannon (differential) entropy functional defines $\mathbf{I}(\mathbf{x};\mathbf{z})$ and $\mathbf{T}(\mathbf{z})$ with Eq.~(\ref{eq_mutual_information}) and Eq.~(\ref{eq_total_corr}), respectively:
\begin{equation}\label{eq_mutual_information}
\begin{split}
\mathbf{I}(\mathbf{x}; \mathbf{z}) & = D_{\text{KL}}\left(p(\mathbf{x},\mathbf{z})\|p(\mathbf{x})p(\mathbf{z})\right) \\
& = \int\int p(\mathbf{x},\mathbf{z})\log\left(\frac{p(\mathbf{x},\mathbf{z})}{p(\mathbf{x})p(\mathbf{z})}\right)d\mathbf{x} d\mathbf{z}
\end{split}
\end{equation}

\begin{equation}\label{eq_total_corr}
\begin{split}
\mathbf{T}(\mathbf{z}) & = D_{\text{KL}}\left(p(\mathbf{z})\|p(z_1)\cdots p(z_d)\right) \\
& = \int\cdots\int p(\mathbf{z})\log\left(\frac{p(\mathbf{z})}{p(z_1)\cdots p(z_d)}\right)dz_1\cdots dz_d
\end{split}
\end{equation}

As can be seen, a precise estimation to the joint distribution 
$p(\mathbf{x},\mathbf{z})$ or $p(z_1,z_2,\cdots,z_d)$ is a prerequisite to obtain a reliable evaluation to both $\mathbf{I}(\mathbf{x};\mathbf{z})$ and $\mathbf{T}(\mathbf{z})$. Unfortunately, the density estimation in high or ultra-high dimensional space is always problematic and impractical. 
Taking the simplest $dSprites$ data set as an example, we have $\mathbf{x}\in\mathbb{R}^{4096}$ and $\mathbf{z}\in\mathbb{R}^{10}$ in the general setup.

To circumvent this issue, we use the recently proposed matrix-based R{\'e}nyi's $\alpha$-order entropy functional~\cite{giraldo2014measures,yu2019multivariate} to estimate $\mathbf{I}(\mathbf{x};\mathbf{z})$ and $\mathbf{T}(\mathbf{z})$. Unlike the standard Shannon entropy functional, the novel estimator is defined over the normalized eigenspectrum of a Hermitian matrix of the projected data in a reproducing kernel Hilbert space (RKHS). In this way, it avoids the explicit estimation of the underlying distributions of data. In the remaining of this section, we only demonstrate our observations related to the fundamental properties of VAEs. We leave the implementation details and the robustness of our estimator with respect to different hyper-parameter setting to the supplementary material.

\subsection{Compression and Positive Correlation}

\begin{figure}
    \centering
    \includegraphics[width=0.9\textwidth]{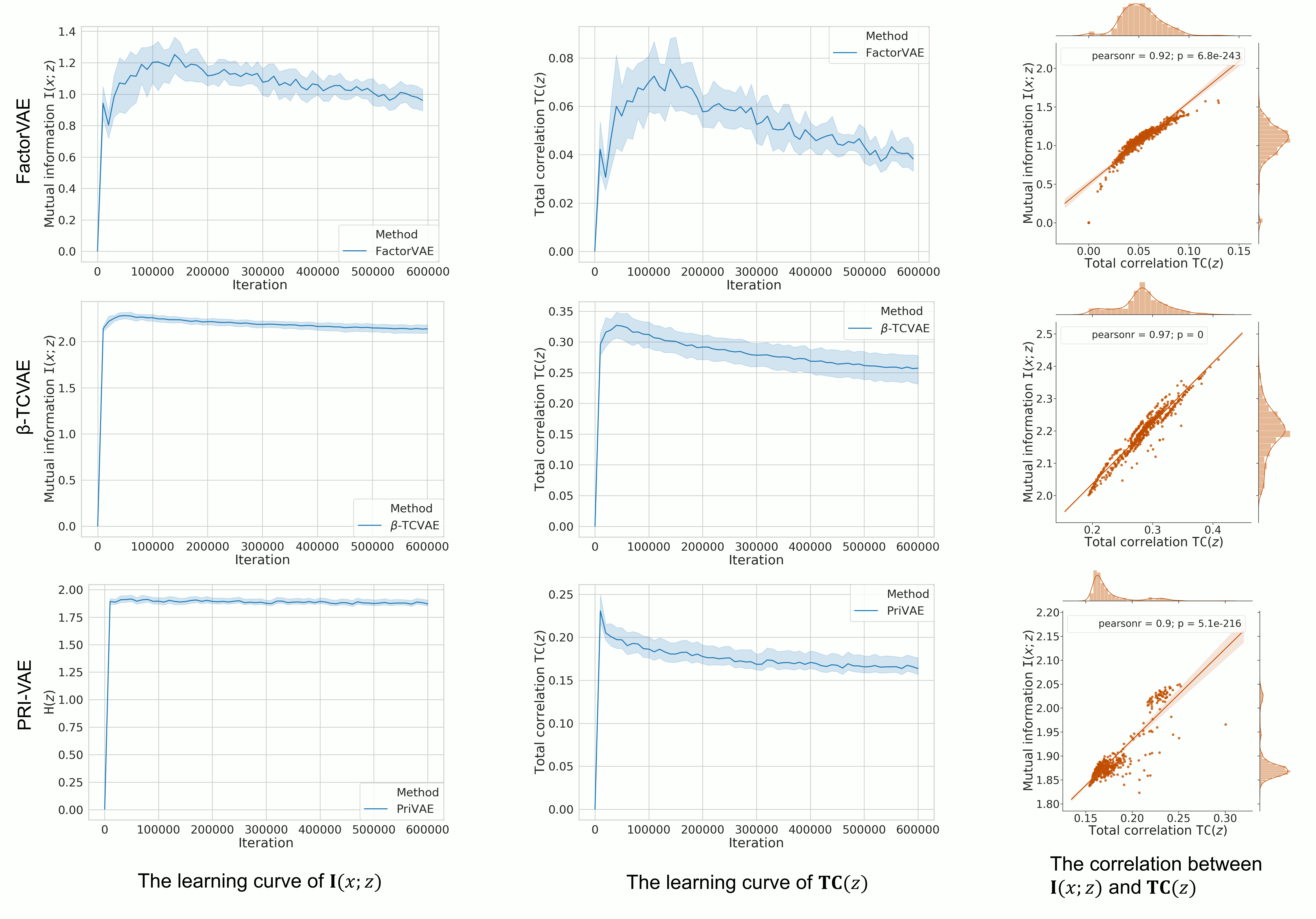}
    \caption{
    The evolution of the value of $\mathbf{I}(\mathbf{x};\mathbf{z})$ (left column) and the value of $\mathbf{T}(\mathbf{z})$ (middle column) across training iterations for FactorVAE, $\beta$-TCVAE and PRI-VAE on \textit{Cars3d} data set. The right column shows the positive correlation between these two values.}
    \label{fig:learning_curve_of_MI_TC_alpha-1_1}
\end{figure}

Fig.~\ref{fig:learning_curve_of_MI_TC_alpha-1_1} demonstrates the evolution of $\mathbf{I}(\mathbf{x};\mathbf{z})$ and $\mathbf{T}(\mathbf{z})$ for two representative VAEs across training iterations. 
Obviously, during the common SGD optimization, both $\mathbf{I}(\mathbf{x};\mathbf{z})$ and $\mathbf{T}(\mathbf{z})$ show two separate phases: an early ``fitting" phase, in which both $\mathbf{I}(\mathbf{x};\mathbf{t})$ and $\mathbf{I}(\mathbf{t};\mathbf{y})$ increase rapidly, and a later ``compression" phase, in which there is a reversal such that $\mathbf{I}(\mathbf{x};\mathbf{t})$ and $\mathbf{I}(\mathbf{t};\mathbf{y})$ continually decrease.
Our observation fits well with the arguments of the mainstream information theory community.



More interestingly, we also observe a significant positive correlation between $\mathbf{I}(\mathbf{x};\mathbf{z})$ and $\mathbf{T}(\mathbf{z})$, as manifested by their Pearson correlation coefficient.
Note that, a small value of $\mathbf{T}(\mathbf{z})$ is always a good indicator to model disentanglement~\cite{chen2018isolating}. 
According to our observation, if $\mathbf{T}(\mathbf{z})$ is optimized to a small value, $\mathbf{I}(\mathbf{x};\mathbf{z})$ should be small accordingly (even zero if fully disentangled). However, it is counter-intuitive that an extremely small value of $\mathbf{I}(\mathbf{x};\mathbf{z})$ is able to faithfully reconstruct the input~\cite{higgins2017beta,chen2018isolating}. \textcolor{black}{In this sense, our observation provides empirical evidence to support the argument in~\cite{zhao2019infovae}, which states that existing VAE models tend to learn uninformative latent representation. It also suggests there is always a trade-off between reconstruction and disentanglement~\cite{lezama2018overcoming,esmaeili2019structured,kim2018disentangling}}. 

\subsection{Observations with Other Estimators}
\begin{figure}
    \centering
    \subfloat{\includegraphics[width=0.5\textwidth]{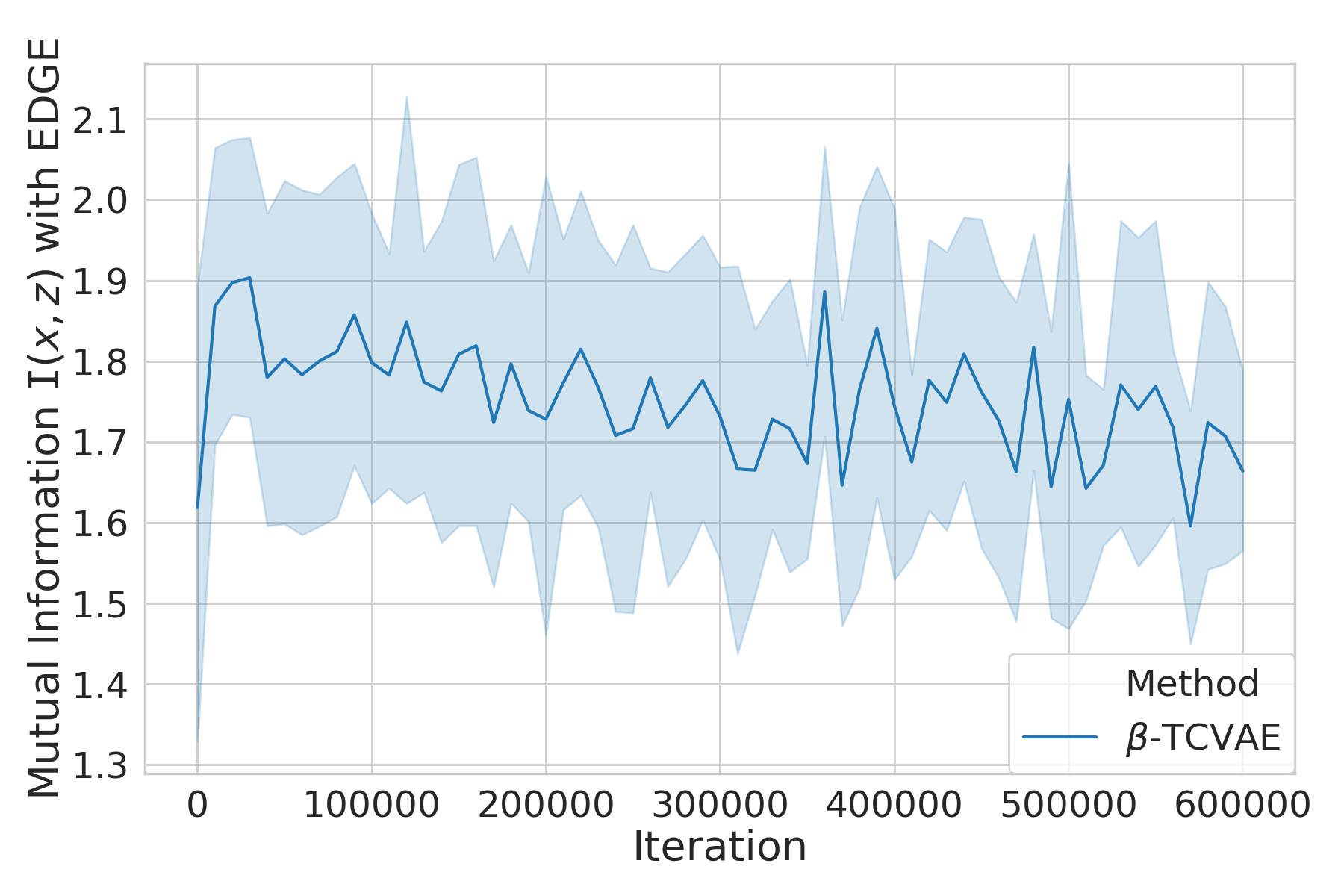}} \hspace{0.05\textwidth}
    \subfloat{\includegraphics[width=0.4\textwidth]{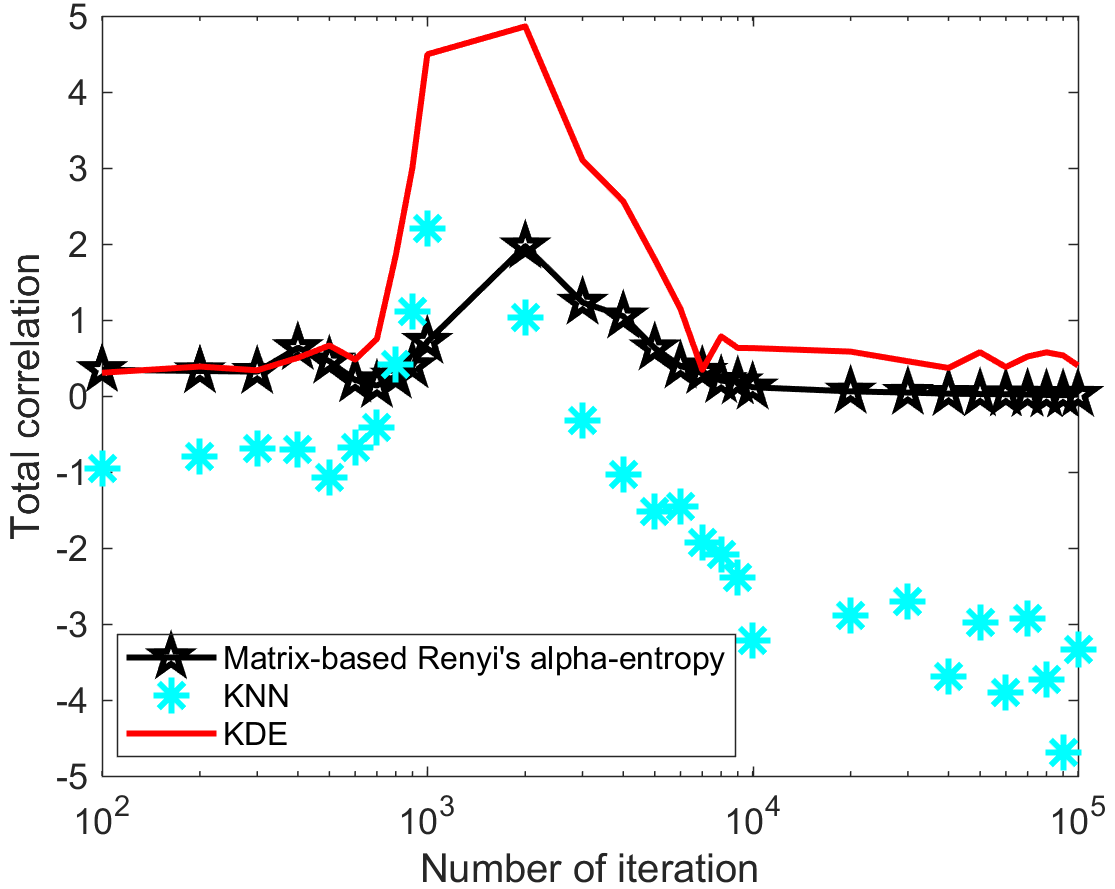}}
    \caption{The evolution of value of $\mathbf{I}(\mathbf{x};\mathbf{z})$ (left) measured by EDGE estimator and $\mathbf{T}(\mathbf{z})$ (right) measured by $k$NN and KDE.} 
    \label{fig:edge}
\end{figure}

Our aim here is to verify if the general observation made by our suggested matrix-based R{\'e}nyi's $\alpha$-entropy functional are consistent with other existing Shannon estimators. To this end, we estimate $\mathbf{T}(\mathbf{z})$ with the widely used $k$NN estimator~\cite{kraskov2004estimating} and kernel density estimator (KDE)~\cite{moon1995estimation}. We also apply the recently proposed  ensemble dependency graph estimator (EDGE)~\cite{noshad2019scalable} to measure $\mathbf{I}(\mathbf{x};\mathbf{z})$. Fig.~\ref{fig:edge} suggests that our general observation is in accordance with other popular estimators. However, note that both $k$NN and KDE suffer from the curse of dimensionality, and it still remains a problem on estimating total correlation with EDGE. 


\section{Experiments}\label{sec:experiment}
We evaluate the performances of our proposed PRI-VAE on four benchmark data sets to demonstrate its capability to learn disentangled, interpretable latent representations, and reconstruct input signal with high fidelity. We compare PRI-VAE with six SOTA VAE models, namely $\beta$-VAE~\cite{higgins2017beta}, AnnealedVAE~\cite{burgess2018understanding}, DIP-VAE~\cite{kumar2018variational}, InfoVAE~\cite{zhao2019infovae}, FactorVAE~\cite{kim2018disentangling}, and $\beta$-TCVAE~\cite{chen2018isolating}.
We evaluate the disentanglement and reconstruction performances of all competing models quantitatively (in Section~\ref{sec:quantitative}) and qualitatively (in Section~\ref{sec:qualitative}). We also illustrate the potential implications of our methodology for future work and generalize the idea of PRI-VAE to a higher level in Section~\ref{sec:implication}.
For a fair comparison, all competing methods are trained with the same network architecture, optimizer, and mini-batch and evaluated by unified disentangling metrics~\cite{locatello2019challenging}. 



\subsection{Quantitative Evaluation} \label{sec:quantitative}
We perform quantitative evaluation on \textit{dSprites}~\cite{dSprites17} and \textit{Cars3D}~\cite{reed2015deep}. Samples in both data sets are generated with ground-truth independent latent factors. 
Since most of the existing metrics on disentanglement evaluation are positively correlated \cite{locatello2019challenging}, we consider the most widely used Mutual Information Gap (MIG)~\cite{chen2018isolating} and DCI Disentanglement metric~\cite{ridgeway2018learning,locatello2019challenging} in this work.




\subsubsection{Comparison of PRI-VAE and InfoVAE Family}\label{sec:PRI_synthetic_comparison1}

The most similar objective to PRI-VAE is the InfoVAE family~\cite{zhao2019infovae} with the regularization term $\alpha\circled{\footnotesize 1} + \lambda\circled{\footnotesize 2}$.
Despite reaching similar objectives, the routes to this result are rather different. 
Our PRI-VAE is motivated by an information-theoretic perspective in the sense that we want to learn a compact (regularized by $\alpha\mathbf{H}(\mathbf{z})$) latent representation that is able to reconstruct input signal (regularized by $\mathbb{E}_{p(\mathbf{x})}\left[\mathbb{E}_{q(\mathbf{z}|\mathbf{x})}\left[\log p(\mathbf{x}|\mathbf{z})\right]\right]$) and is also disentangled (regularized by $\beta D_{KL}[q_{\phi}(\mathbf{z})\| p(\mathbf{z})]$). By contrast, InfoVAE family originates from a generative model perspective that attempts to address two fundamental issues in the original VAE (i.e., the inaccurate amortized inference distributions and the vanishing relevance of latent variable and input signal). 
\textcolor{black}{Although authors of InfoVAE suggest setting $\alpha=0$ to completely drop \circled{\footnotesize 1}, we demonstrate here the superiority gained from the difference, i.e.,  $\alpha(\mathbf{H}(\mathbf{z}) - \mathbf{I}(\mathbf{x};\mathbf{z}))$.} 


\begin{figure}[t]
    \centering
    \subfloat{\includegraphics[width=0.45\textwidth]{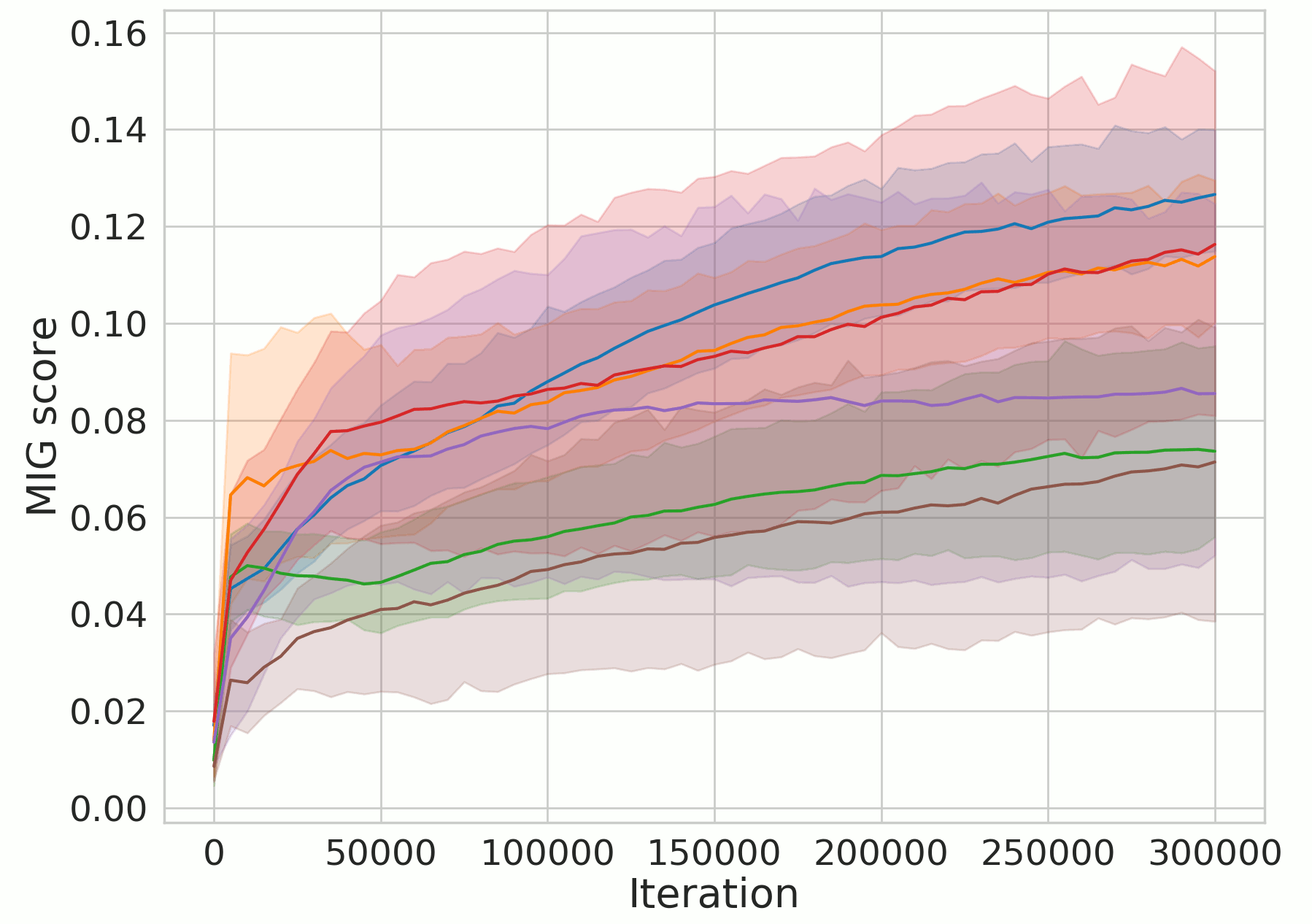}} \hspace{0.03\textwidth}
    \subfloat{\includegraphics[width=0.47\textwidth]{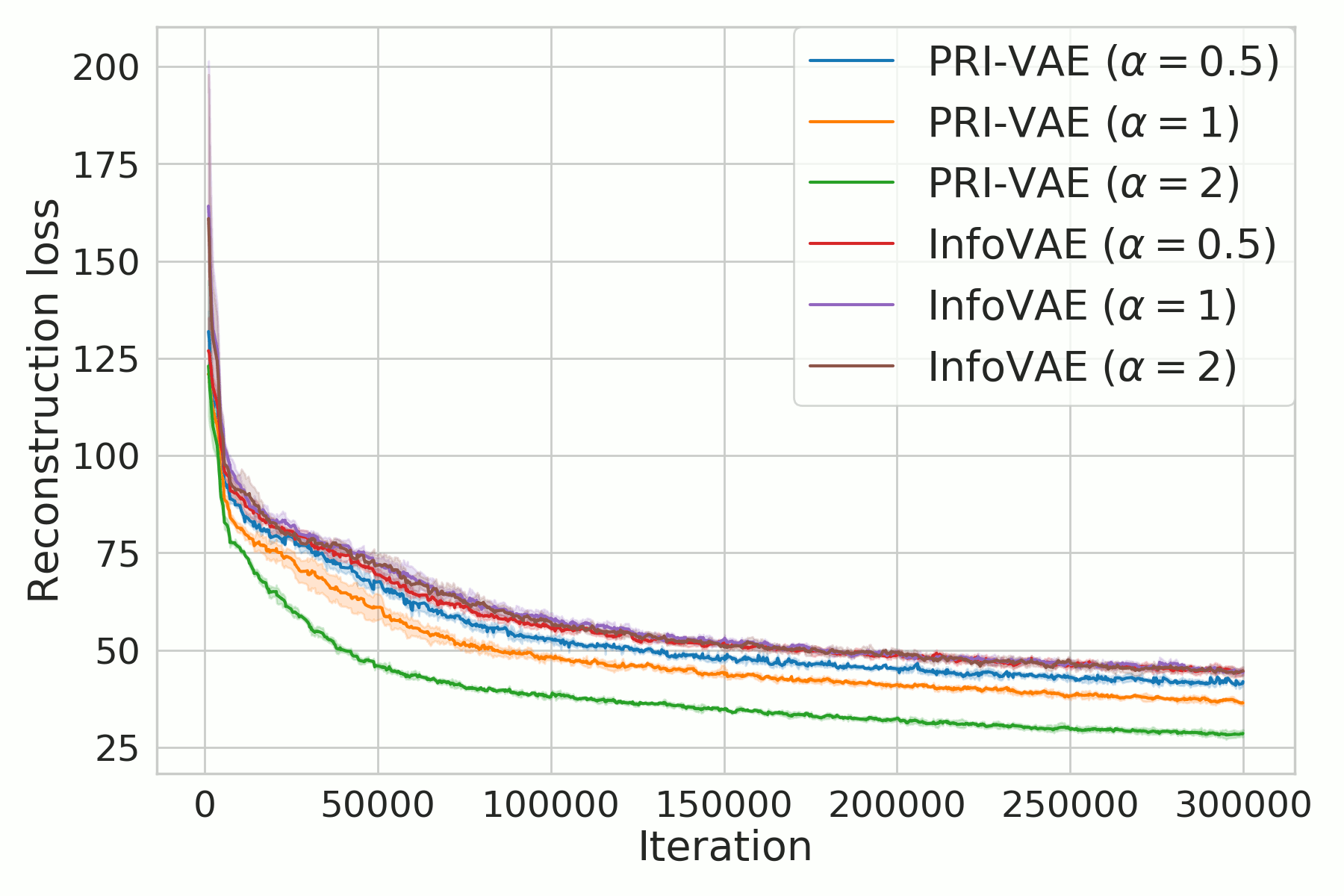}} 
    \caption{The evolutions of MIG score (left) and reconstruction loss (right, after exponential moving average with weight $0.8$) for PRI-VAE and InfoVAE family with respect to different values of $\alpha$ when $\beta$ is fixed to $4$.} 
    \label{fig:comparison_infovae_dSprites}
    \vspace{-1em}
\end{figure}

\begin{figure}
    \centering
    \subfloat{\includegraphics[width=0.45\textwidth]{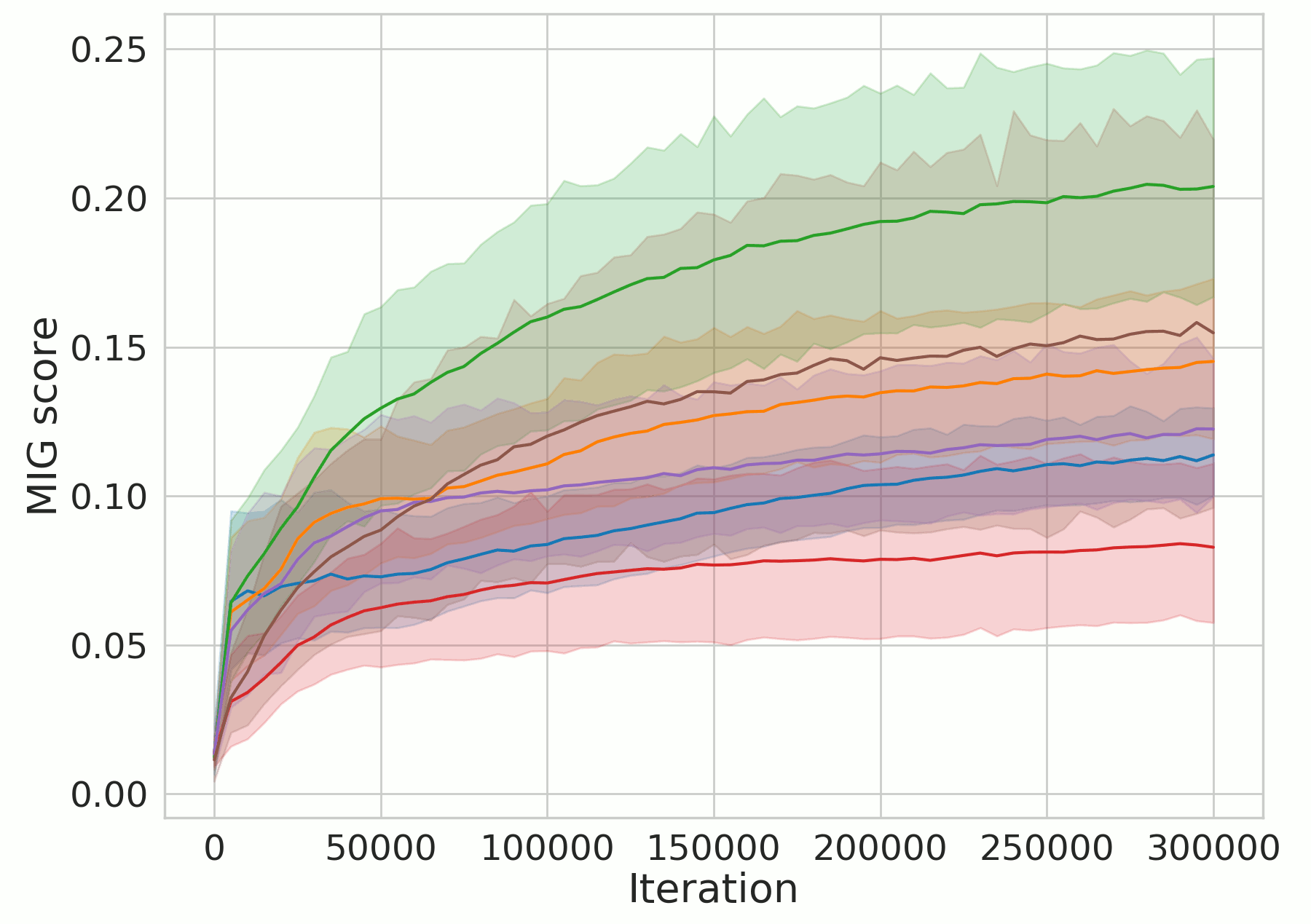}} \hspace{0.03\textwidth}
    \subfloat{\includegraphics[width=0.47\textwidth]{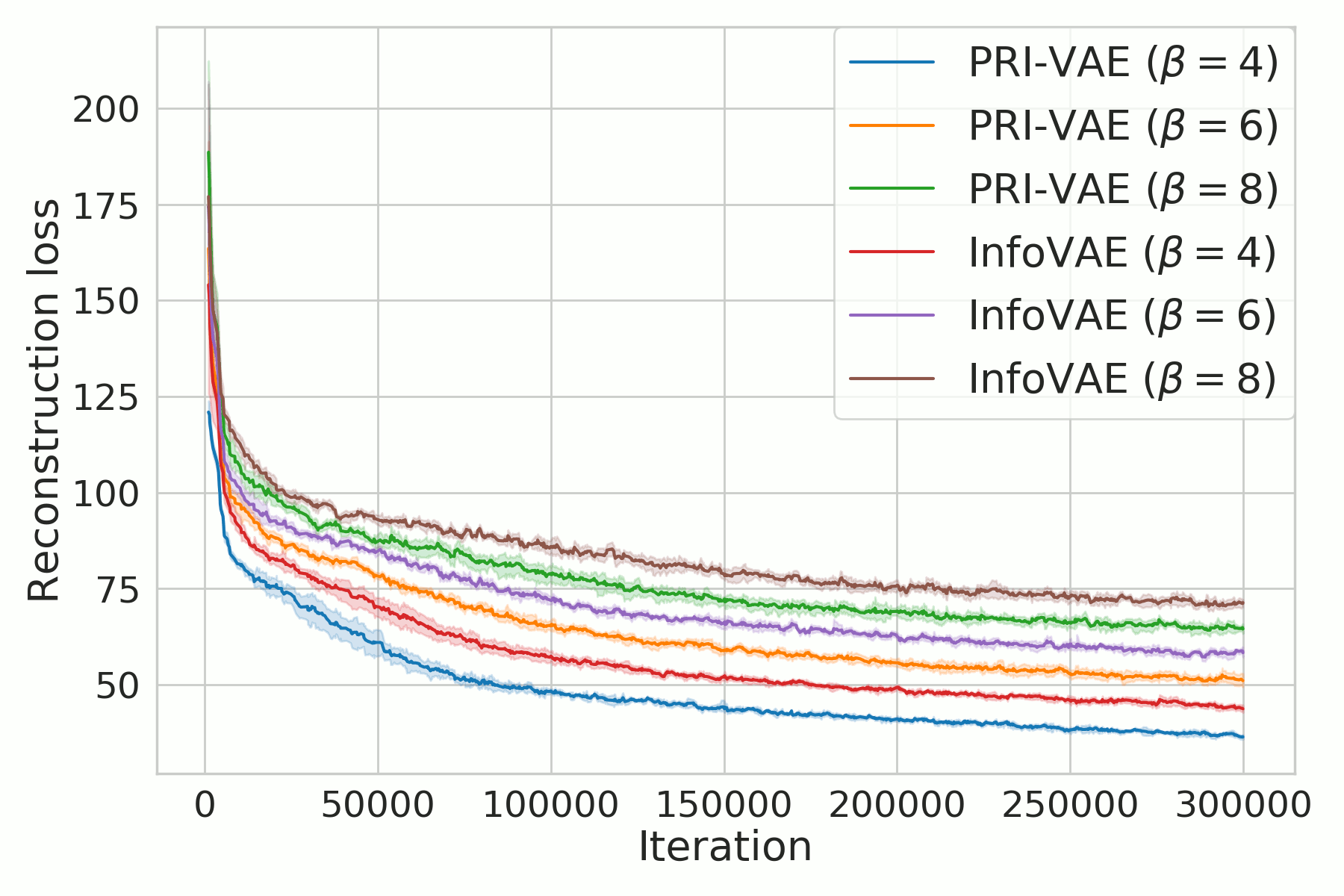}} 
    \caption{The evolutions of MIG score (left) and reconstruction loss (right, after exponential moving average with weight $0.8$) for PRI-VAE and InfoVAE family with respect to different values of $\beta$ when $\alpha$ is fixed to $1$.} 
    \label{fig:comparison_infovae_2_dSprites}
\end{figure}


We first evaluate the performances of PRI-VAE and InfoVAE family with respect to different values of $\alpha$ by fixing $\beta=4$ (a common setting in previous works~\cite{higgins2017beta,esmaeili2019structured}).
Fig.~\ref{fig:comparison_infovae_dSprites} demonstrates the evolution of the MIG score and the reconstruction loss across training iterations on \textit{dSprites} data set. Under the same hyper-parameter setting (i.e., same value of $\alpha$), PRI-VAE can quickly obtain more disentangled latent representation with smaller reconstruction error than its InfoVAE counterpart. This is not surprising. Note that $\mathbf{H}(\mathbf{z}) - \mathbf{I}(\mathbf{x};\mathbf{z})\geq0$, i.e., $\mathbf{H}(\mathbf{z})$ is always a heavy penalty than $\mathbf{I}(\mathbf{x};\mathbf{z})$. Therefore, a faster convergence is expected. 




Fig.~\ref{fig:comparison_infovae_2_dSprites} reaffirms the advantage of PRI-VAE over InfoVAE family concerning different values of $\beta$. Another interesting observation is that, with the increase of $\beta$, the MIG score keeps increasing, whereas the reconstruction quality is reduced. This is probably because there is a trade-off between reconstruction and disentanglement~\cite{lezama2018overcoming,kim2018disentangling,esmaeili2019structured}. 



\subsubsection{Comparison of PRI-VAE and other SOTA VAE Models} \label{sec:PRI_synthetic_comparison2}
Next, we compare PRI-VAE with VAE, $\beta$-VAE, AnnealedVAE, DIP-VAE-\RNum{1}, DIP-VAE-\RNum{2}, FactorVAE and $\beta$-TCVAE.
As recommended in~\cite{locatello2019challenging}, we finalize the hyperparameters for each methods by sweeping over a wide enough regularization weight range. We then fix the hyperparameters of all methods on different data sets.
Specifically, we select $\beta=4$ for $\beta$-VAE and $\beta$-TCVAE, $\gamma=30$ for FactorVAE, $\lambda_{od}=5$, $\lambda_{d}=10\lambda_{od}$ for DIP-VAE-\RNum{1} and $\lambda_{od}=5$, $\lambda_{d}=\lambda_{od}$ for DIP-VAE-\RNum{2}.
For AnnealedVAE, we fix $\gamma=1000$ and linearly increase the channel capacity $C$ from $0$ to $25$ over the course of $10,000$ training steps.
For our model, we select $\alpha=0.6$ and $\beta=6.0$\footnote{\textcolor{black}{We recommend $\beta\geq 2\alpha$. This is because $\mathbf{z}$ converges to a single point if $\beta\leq \alpha$ (see Fig.~\ref{fig:pri_demo}), from which it is hard to reconstruct $\mathbf{x}$.}}. Ten independent runs are performed.

\begin{figure}[t]
    \centering
    \subfloat{\includegraphics[width=0.44\textwidth]{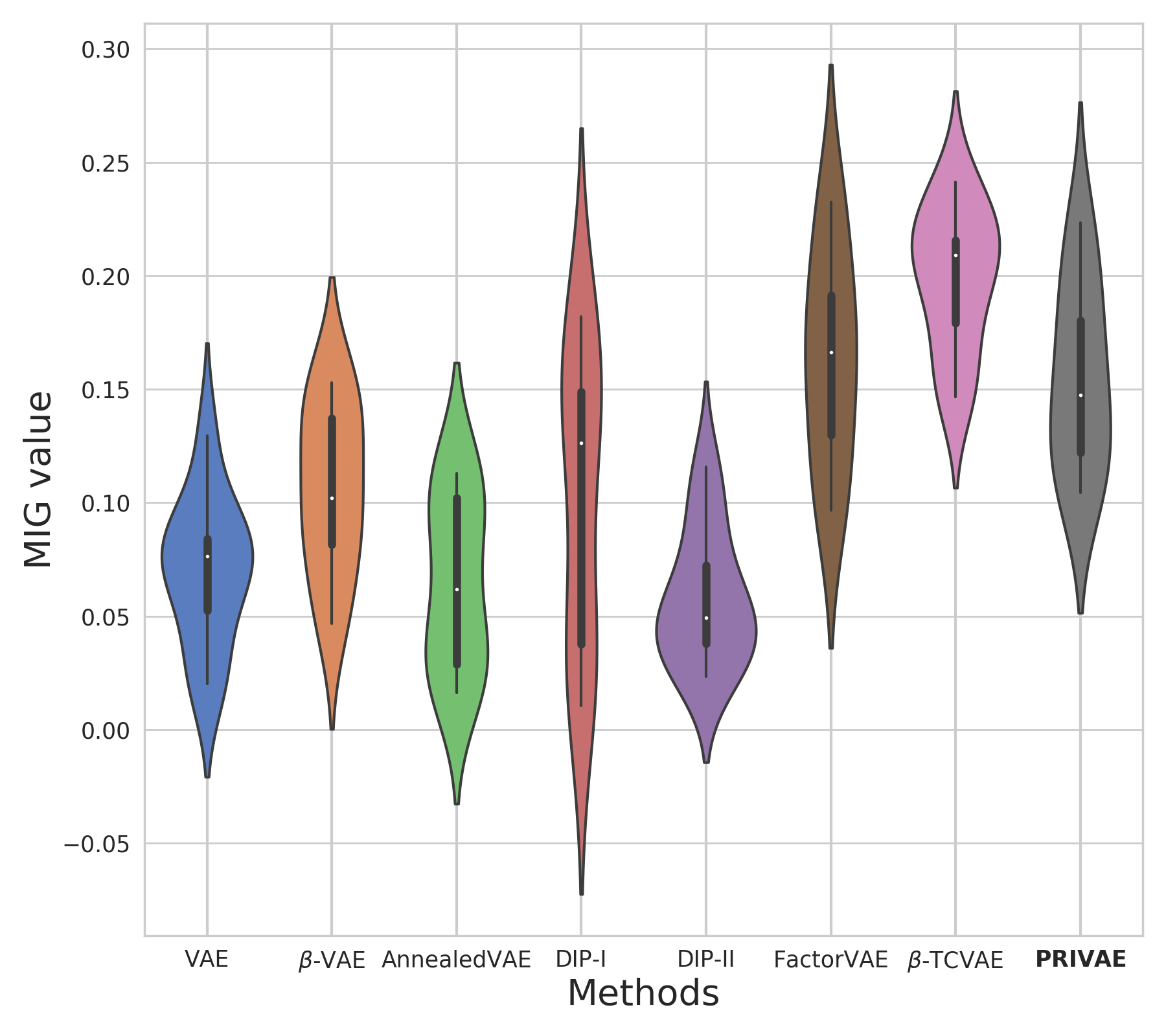}} \hspace{0.05\textwidth}
    \subfloat{\includegraphics[width=0.44\textwidth]{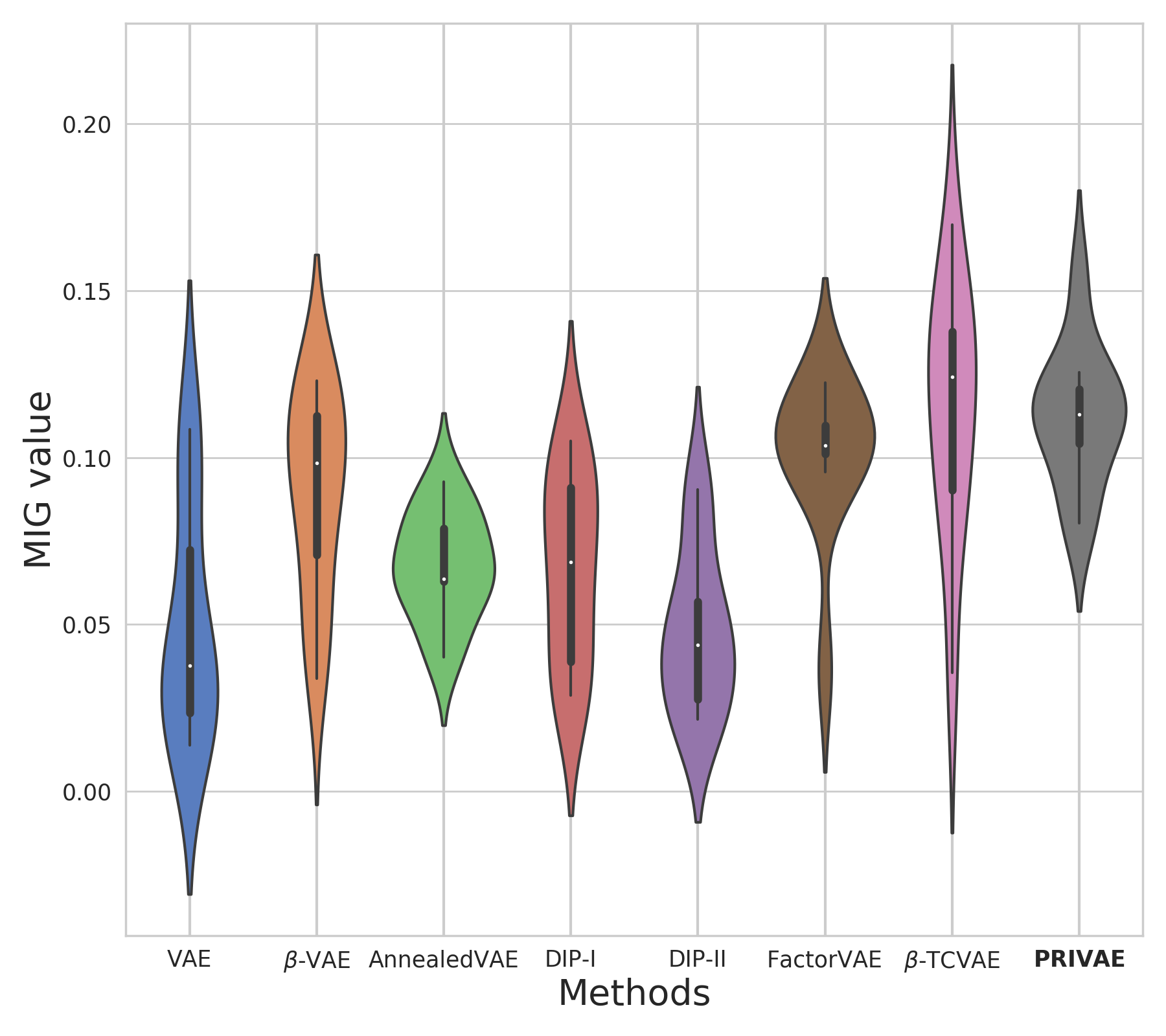}} 
    \caption{The MIG score comparison on \textit{dSprites} (left) and \textit{Cars3D} (right) data sets.} 
    \label{fig:compare_baseline_mig}
\end{figure}

Fig.~\ref{fig:compare_baseline_mig} demonstrates the MIG score for all competing models in \textit{dSprites} and \textit{Cars3D}, respectively\footnote{Same to \cite{locatello2019challenging,esmaeili2019structured}, we also observe that the hyper-parameter tuning and the random seed have substantial impacts to the performances of different models.}. 
The comparison result on DCI disentanglement metric is similar, which is shown in Appendix Fig.~\ref{fig:compare_baseline_dci}.
As can be seen, PRI-VAE achieves consistent better disentanglement performance than VAE, $\beta$-VAE, AnnealedVAE, and DPI-VAE, and preserves a satisfactory reconstruction quality suggested by our visualization results (see Appendix Fig.~\ref{fig:privae_reconstruction}).
However, $\beta$-TCVAE and FactorVAE  obtain higher disentanglement scores than ours. 
This is not surprising. Both FactorVAE and $\beta$-TCVAE add a new regularization term $\mathbf{T}(\mathbf{z})$ with extra weight to the original VAE objective. 
One should note that $\mathbf{T}(\mathbf{z})$ is always a more explicit and heavy penalty than $D_{KL}(q_\phi(\mathbf{z})\|p(\mathbf{z}))$ to encourage independence of each dimension of $\mathbf{z}$~\cite{chen2018isolating}.
We will show in Section~\ref{sec:implication_2} that the merits of PRI-VAE and $\beta$-TCVAE can be merged to significantly improve the performances of both methods.  
 
\subsection{Qualitative Evaluation} \label{sec:qualitative}

\begin{figure}[h]
\vspace{-0em}
    \centering
    \includegraphics[width=0.8\textwidth]{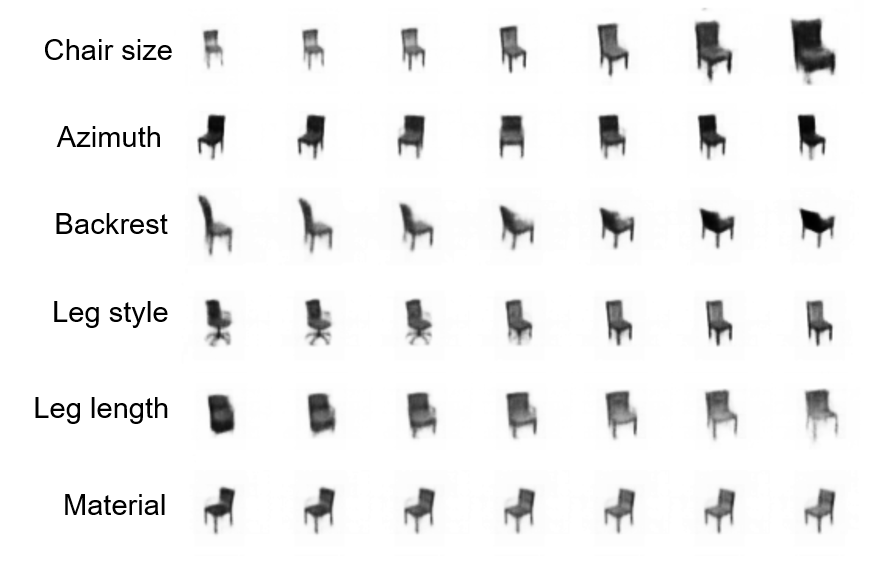}
    \vspace{-1em}
    \caption{Latent traversals for the PRI-VAE ($\alpha=0.6, \beta=4$) trained on \textit{3D chairs}. Each column corresponds to varying a single latent unit. The traversal is over the $[-2, 2]$ range.}
    \label{fig:chairs_traversal_partial}
\end{figure}

\begin{figure}[h]
    \centering
    \includegraphics[width=0.78\textwidth]{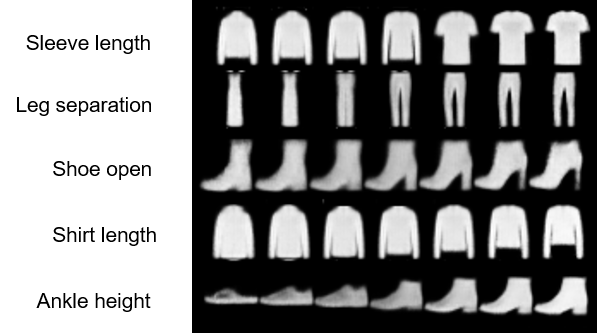}
    \caption{Latent traversals for the PRI-VAE ($\alpha=0.6, \beta=8$) trained on \textit{Fashion-MNIST}. Each column corresponds to varying a single latent unit. The traversal is over the $[-3, 3]$ range.}
    \label{fig:FashionMNIST}
\end{figure}

We then quantitatively evaluate PRI-VAE on \textit{3D Chairs}~\cite{aubry2014seeing}, and \textit{Fashion-MNIST}~\cite{xiao2017fashion}.
Fig.~\ref{fig:chairs_traversal_partial} and Fig.~\ref{fig:FashionMNIST} show the traversal results of the latent variables on both data sets, respectively. Similar to $\beta$-TCVAE, PRI-VAE is able to discover six factors of variation in \textit{3D Chairs}, such as chair size, azimuth, backrest, leg style, leg length, and chair material. By contrast, $\beta$-VAE can only learn four factors~\cite{chen2018isolating}. Moreover, PRI-VAE discovers a unique factor (i.e., leg separation) in~\textit{Fashion-MNIST}, which has never been mentioned in previous works. 

\vspace{-1em}
\subsection{Implications for Future Work} \label{sec:implication}

We finally present two positive implications of our methodology for future work with convincing empirical validation.

\subsubsection{Merging the Merits of PRI-VAE and $\beta$-TCVAE}\label{sec:implication_2}

We continue our discussion in Section~\ref{sec:PRI_synthetic_comparison2} to demonstrate the benefits gained from merging the merits of PRI-VAE and $\beta$-TCVAE.
As discussed earlier, FactorVAE and $\beta$-TCVAE have better disentanglement performance than others by explicitly regularizing an extra term $\mathbf{T}(\mathbf{z})$. 
On the other hand, the PRI is geometrically interpretable and enables faster convergence. 
Therefore, a natural idea to combine the merits of both methodologies is via the following objective:
\textcolor{black}{
\begin{equation}
\begin{split}
&\mathcal{L}_{\mathrm{PRI}-\mathrm{VAE}^\star} \\
    &=\mathbb{E}_{p(\mathbf{x})}\left[\mathbb{E}_{q(\mathbf{z} | \mathbf{x})}\left[\log p(\mathbf{x} | \mathbf{z})\right]\right] - \alpha H(\mathbf{z}) \\
    &-\beta D_{KL}[q(\mathbf{z}) \| p(\mathbf{z})] - \gamma D_{KL}(q(\mathbf{z}) \| \prod_{j} q\left(\mathbf{z}_{j}\right)).
\end{split}
\label{eq:pri_vae_star}
\end{equation}}

We term Eq.~(\ref{eq:pri_vae_star}) PRI-VAE$^\star$ since it just adds a weighted term $\gamma\mathbf{T}(\mathbf{z})$ to the original PRI-VAE objective.
\textcolor{black}{In Eq.~(\ref{eq:pri_vae_star}), the term $\alpha H(\mathbf{z}) + \beta D_{KL}q(\mathbf{z})||p(\mathbf{z}))$ can be interpreted as an upper bound of the exact information rate $\mathbf{I}(\mathbf{x};\mathbf{z})$, whereas the new regularization term $\gamma D_{KL}(q(\mathbf{z}) \| \prod_{j} q\left(\mathbf{z}_{j}\right))$ further enforces the disentanglement. This way, we are actually optimizing a regularized rate-distortion problem. Given that minimizing $\mathbf{T}(\mathbf{z})$ is likely to incur the decrease of $\mathbf{I}(\mathbf{x};\mathbf{z})$, the distortion may increase accordingly. }

We compare PRI-VAE$^\star$ with PRI-VAE and $\beta$-TCVAE on \textit{dSprites} and \textit{Cars3D}.
We keep the same hyper-parameter setting for PRI-VAE and $\beta$-TCVAE as shown in Section~\ref{sec:PRI_synthetic_comparison2}. 
For PRI-VAE$^\star$, we set $\alpha=0.5, \beta=1.0$ and $\gamma=4.0$. 
Table~\ref{tb:pri_vae_star} summarizes the quantitative evaluation results and DCI comparison is shown in Appendix Table~\ref{tb:pri_vae_star_dci}.
As can be seen, PRI-VAE$^\star$ has improved disentanglement than PRI-VAE and it is slightly better than $\beta$-TCVAE. 
Note that the hyperparameters of all the methods are not additionally tuned.

\begin{table}[h]
\caption{The MIG values of $\beta$-TCVAE, PRI-VAE and PRI-VAE$^\star$ on \textit{dSprites} and \textit{Cars3D}. We report the mean value over $10$ runs and the standard deviation in parentheses.} \label{tb:pri_vae_star}
\renewcommand{\arraystretch}{1.5}
\begin{center}
\begin{tabular}{llll}
\hline
 & $\beta$-TCVAE & PRI-VAE & PRI-VAE$^\star$ \\ \hline
\textit{dSprites} & 0.194 (0.030) & 0.155 (0.042) & \textbf{0.198 (0.063)} \\ \hline
\textit{Cars3D} & 0.113 (0.038) & 0.112 (0.021) & \textbf{0.132 (0.033)} \\ \hline
\end{tabular}
\end{center}
\end{table}

\subsubsection{Replacing $\mathbf{I}(\mathbf{x};\mathbf{z})$ with $\mathbf{H}(\mathbf{z})$ in SOTA VAE models}

\begin{figure}[hb]
    \centering
    \subfloat{\includegraphics[width=0.44\textwidth]{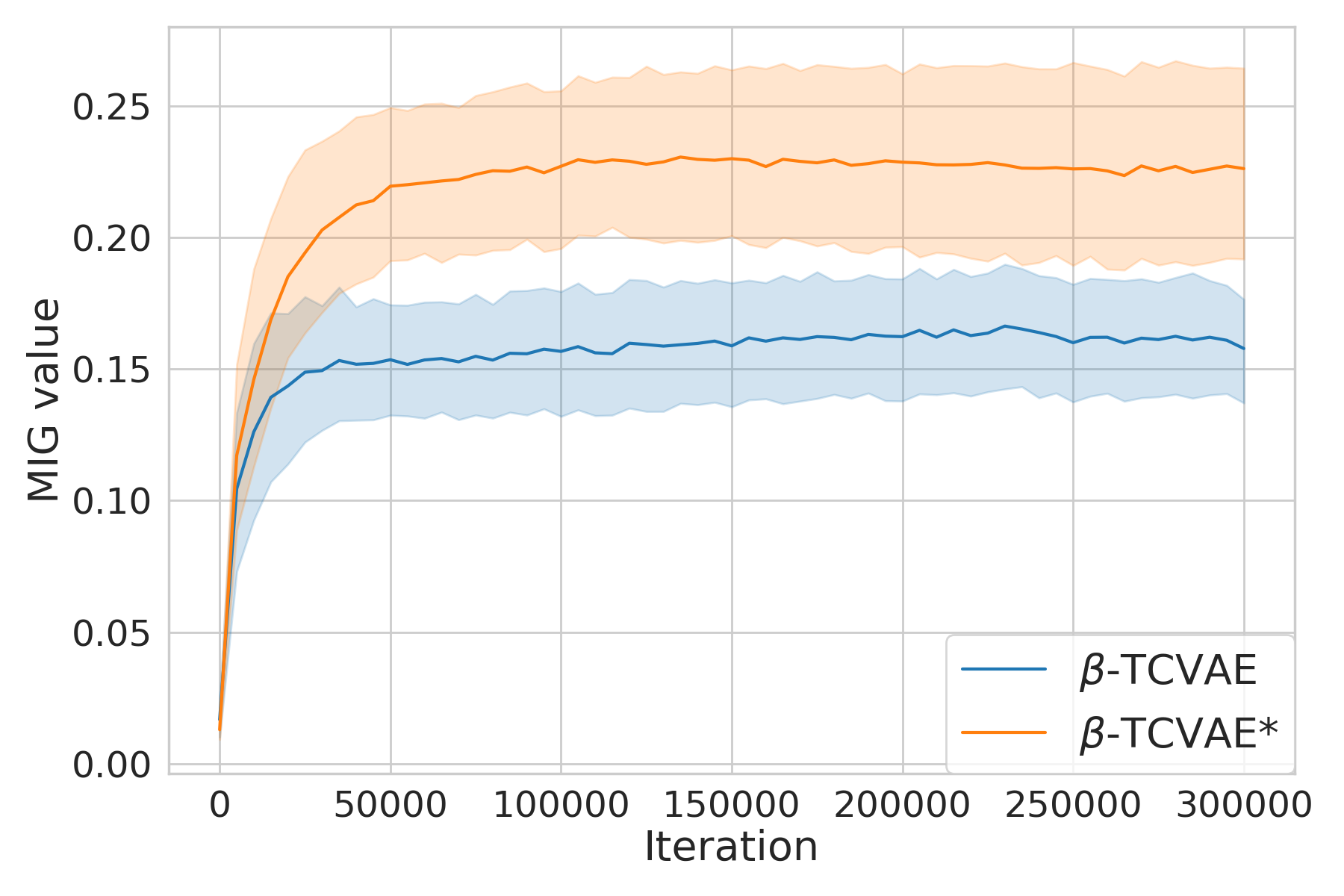}} \hspace{0.03\textwidth}
    \subfloat{\includegraphics[width=0.44\textwidth]{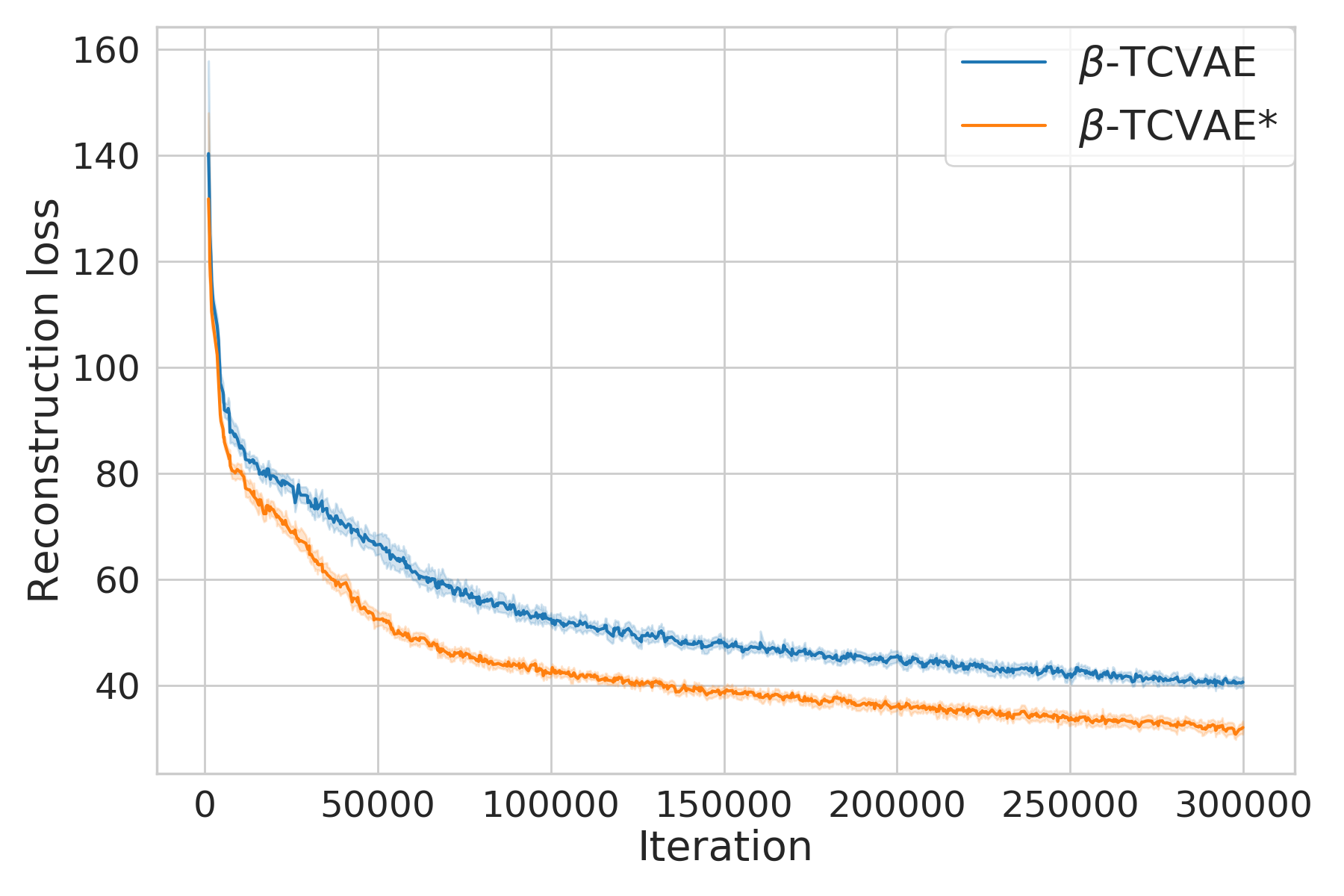}} 
    \caption{The comparison of disentanglement (left) and reconstruction (right) between $\beta$-TCVAE$^\star$ and $\beta$-TCVAE on \textit{dSprites}.} 
    \label{fig:pri_beta_tcvae}
\end{figure}

As demonstrated in Section~\ref{sec:PRI_synthetic_comparison1}, PRI-VAE achieves an obvious performance gain over InfoVAE family by replacing $\mathbf{I}(\mathbf{x};\mathbf{z})$ with $\mathbf{H}(\mathbf{z})$. 
We now investigate if this modification can be generalized to other SOTA VAE models. 
Again, we use the $\beta$-TCVAE as an example, and suggest the following objective (term it $\beta$-TCVAE$^\star$):
{\small
\begin{equation}
\begin{split}
&\mathcal{L}_{\beta-\mathrm{TCVAE}^\star} \\
    &=\mathbb{E}_{p(\mathbf{x})}\left[\mathbb{E}_{q(\mathbf{z} | \mathbf{x})}\left[\log p(\mathbf{x} | \mathbf{z})\right]\right] - H(\mathbf{z}) \\
    &-\beta D_{KL}(q(\mathbf{z}) \| \prod_{j} q\left(\mathbf{z}_{j}\right)) - \sum_{j} D_{KL}(q\left(\mathbf{z}_{j}\right) \| p\left(\mathbf{z}_{j}\right)).
\end{split}
\end{equation}
}
We compare $\beta$-TCVAE$^\star$ with the original $\beta$-TCVAE with the same value of $\beta$. The comparison result on \textit{dSprites} is shown in Fig.~\ref{fig:pri_beta_tcvae}.
It is interesting to find that $\beta$-TCVAE$^\star$ is consistently better than classical $\beta$-TCVAE in terms of both disentanglement and reconstruction. 



\section{Related Work}
\subsection{Disentangled Representation Learning with VAEs}
Before our work, many efforts have been made for disentangled representation learning with the VAE framework. 
In Section~\ref{sec:PRI_VAE_relation} we illustrate the connections between our PRI-VAE and some prevalent VAEs.
More recently, HFVAE~\cite{esmaeili2019structured} decomposes the VAE objective into four terms, which provides a unified view to most VAE variants.
Interested readers can refer to~\cite{tschannen2018recent} for a comprehensive survey.

\vspace{-1em}
\subsection{Interpreting DNNs with Information Theory}
There has been a growing interest in understanding DNNs using information theory. According to~\cite{shwartz2017opening}, a DNN can be analyzed by measuring the amount of information that each hidden layer's representation $\mathbf{t}$ preserves about the input signal $\mathbf{x}$ with respect to the desired response $\mathbf{y}$ (i.e., $\mathbf{I}(\mathbf{x};\mathbf{t})$ with respect to $\mathbf{I}(\mathbf{t};\mathbf{y})$). 
This technique has been applied to various DNN architectures, such as the the multilayer perceptrons (e.g.,~\cite{chelombiev2019adaptive,shwartz2017opening}), the deterministic autoencoders (e.g.,~\cite{yu2019understanding}) and the convolutional neural networks (e.g.,~\cite{noshad2019scalable,yu2018understanding}).
In general, most of these recent studies suggest that both $\mathbf{I}(\mathbf{x};\mathbf{t})$ and $\mathbf{I}(\mathbf{t};\mathbf{y})$ undergo separate ``fitting'' and ``compression'' phases, which is consistent with our observations as described in Section~\ref{sec:Empirical_behavior}.

Before our work,~\cite{burgess2018understanding} interpreted the emergence of disentangled representation in $\beta$-VAE with the rate-distortion (RD) theorem~\cite[Chapter~10]{cover2012elements}. Similarly,~\cite{alemi2018information} uses the RD function to explore the Pareto front of existing VAEs in terms of both rate (i.e., the average number of additional nats necessary to encode samples from the encoder) and distortion (i.e., the reconstruction error). On the other hand,~\cite{achille2018information} illuminates the connection between VAE and IB approach.

\section{Conclusion}

In this paper, we introduced the recently proposed matrix-based R{\'e}nyi's $\alpha$-entropy functional to interpret the dynamics of learning of VAE models. We also developed a novel VAE learning objective and establish its connections to prior art. Extensive experimental results on four benchmark data sets suggest that our objective can obtain more interpretable factors and comparable disentangled representation while preserving a high reconstruction fidelity. Moreover, our objective can be easily tailored to expand its capacity; and the merit of our objective can be generalized to other SOTA models.




\bibliographystyle{ieeetrans}
\bibliography{Refs}

%








\clearpage


%

\setcounter{table}{0}
\renewcommand{\thetable}{\Alph{section}\arabic{table}}
\setcounter{figure}{0}
\renewcommand{\thefigure}{\Alph{section}\arabic{figure}}
\appendices
\section{Proofs}
\subsection{The Decomposition of $D_{KL}(q_\phi(\mathbf{z}|\mathbf{x})\|p(\mathbf{z}))$}
Suppose the joint distribution of the data and the encoding distribution is given by $q(\mathbf{z},\mathbf{x})=p(\mathbf{x})q_\phi(\mathbf{z}|\mathbf{x})$, the KL term $D_{KL}(q_\phi(\mathbf{z}|\mathbf{x})\|p(\mathbf{z}))$ in the original VAE objective can be decomposed as follows~\cite{kim2018disentangling}.
\begin{equation}
\begin{split}
    & \mathbb{E}_{p(\mathbf{x})}[D_{KL}(q_\phi(\mathbf{z}|\mathbf{x})\|p(\mathbf{z}))] \\ 
    & = \mathbb{E}_{p(\mathbf{x})}\mathbb{E}_{q_\phi(\mathbf{z}|\mathbf{x})}[\log\frac{q_\phi(\mathbf{z}|\mathbf{x})}{p(\mathbf{z})}] \\
    & = \mathbb{E}_{q(\mathbf{z}, \mathbf{x})}[\log\frac{q_\phi(\mathbf{z}|\mathbf{x})}{p(\mathbf{z})}\frac{q(\mathbf{z})}{q(\mathbf{z})}] \\
    & = \mathbb{E}_{q(\mathbf{z}, \mathbf{x})}[\log\frac{q_\phi(\mathbf{z}|\mathbf{x})}{q(\mathbf{z})}] + \mathbb{E}_{q(\mathbf{z}, \mathbf{x})}[\log\frac{q(\mathbf{z})}{p(\mathbf{z})}] \\
    & = \mathbb{E}_{q(\mathbf{z}, \mathbf{x})}[\log\frac{q(\mathbf{z}, \mathbf{x})}{q(\mathbf{z})p(\mathbf{x})}] + \mathbb{E}_{q(\mathbf{z})}[\log\frac{q(\mathbf{z})}{p(\mathbf{z})}] \\
    & = \mathbf{I}(\mathbf{x};\mathbf{z})+D_{KL}(q_\phi(\mathbf{z})\|p(\mathbf{z})).
\end{split}
\end{equation}

\section{Experimental Details}
\subsection{Information-Theoretic Quantities Estimation}

Given input $\mathbf{x}\in\mathbb{R}^p$ and latent representation $\mathbf{z}\in\mathbb{R}^d$, the standard Shannon (differential) entropy functional defines $\mathbf{I}(\mathbf{x};\mathbf{z})$ and $\mathbf{T}(\mathbf{z})$ with Eq.~(\ref{eq_mutual_information_supp}) and Eq.~(\ref{eq_total_corr_supp}), respectively:
\begin{equation}\label{eq_mutual_information_supp}
\begin{split}
\mathbf{I}(\mathbf{x}; \mathbf{z}) & = D_{\text{KL}}\left(p(\mathbf{x},\mathbf{z})\|p(\mathbf{x})p(\mathbf{z})\right) \\
& = \int\int p(\mathbf{x},\mathbf{z})\log\left(\frac{p(\mathbf{x},\mathbf{z})}{p(\mathbf{x})p(\mathbf{z})}\right)d\mathbf{x} d\mathbf{z} \\
& = \mathbf{H}(\mathbf{x}) + \mathbf{H}(\mathbf{z}) - \mathbf{H}(\mathbf{x},\mathbf{z}),
\end{split}
\end{equation}

\begin{equation}\label{eq_total_corr_supp}
\begin{split}
\mathbf{T}(\mathbf{z}) & = D_{\text{KL}}\left(p(\mathbf{z})\|p(z_1)\cdots p(z_d)\right) \\
& = \int\cdots\int p(\mathbf{z})\log\left(\frac{p(\mathbf{z})}{p(z_1)\cdots p(z_d)}\right)dz_1\cdots dz_d \\
& = \sum_{i=1}^d \mathbf{H}(z_i) - \mathbf{H}(\mathbf{z}),
\end{split}
\end{equation}
where $\mathbf{H}$ denotes the entropy of a single variable or joint entropy of multiple variables. 

In the following of this section, we first give the novel definitions on the matrix-based R{\'e}nyi's $\alpha$-order entropy and joint entropy, we then elaborate the implementation details in drawing the information plane of VAEs. 

\begin{definition}\cite{giraldo2014measures}
Let $\kappa:\mathcal{X}\times\mathcal{X}\mapsto\mathbb{R}$ be a real valued positive definite kernel that is also infinitely divisible~\cite{bhatia2006infinitely}. Given $X=\{\mathbf{x}^1,\mathbf{x}^2,\cdots,\mathbf{x}^N\}$ and the Gram matrix $K$ obtained from evaluating a positive definite kernel $\kappa$ on all pairs of exemplars, that is $(K)_{ij}=\kappa(\mathbf{x}^i,\mathbf{x}^j)$, a matrix-based analogue to R{\'e}nyi's $\alpha$-entropy for a normalized positive definite (NPD) matrix $A$ of size $n\times n$,  such that $\mathrm{tr}(A)=1$, can be given by the following functional:
\begin{equation}
\mathbf{S}_\alpha(A)=\frac{1}{1-\alpha}\log_2\left(\mathrm{tr}(A^\alpha)\right)=
\frac{1}{1-\alpha}\log_2\big[\sum_{i=1}^N\lambda_i(A)^\alpha\big]
\end{equation}
where $A_{ij}=\frac{1}{N}\frac{K_{ij}}{\sqrt{K_{ii}K_{jj}}}$ and $\lambda_i(A)$ denotes the $i$-th eigenvalue of $A$.
\end{definition}

\begin{definition}\cite{yu2019multivariate}
Given a collection of $N$ samples $\{s_i=(\mathbf{x}_1^i,\mathbf{x}_2^i,\cdots, \mathbf{x}_d^i)\}_{i=1}^N$, where the superscript $i$ denotes the sample index, each sample contains $d$ ($d\geq2$) measurements $\mathbf{x}_1\in \mathcal{X}_1$, $\mathbf{x}_2\in \mathcal{X}_2$, $\cdots$, $\mathbf{x}_d\in \mathcal{X}_d$ obtained from the same realization, and the positive definite kernels $\kappa_1:\mathcal{X}_1\times \mathcal{X}_1\mapsto\mathbb{R}$, $\kappa_2:\mathcal{X}_2\times \mathcal{X}_2\mapsto\mathbb{R}$, $\cdots$, $\kappa_d:\mathcal{X}_d\times \mathcal{X}_d\mapsto\mathbb{R}$, a matrix-based analogue to R{\'e}nyi's $\alpha$-order joint-entropy among $d$ variables can be defined as:
\begin{equation}
\mathbf{S}_\alpha(A_1,A_2,\cdots,A_d)=\mathbf{S}_\alpha\left(\frac{A_1\circ A_2\circ\cdots\circ A_d}{\mathrm{tr}(A_1\circ A_2\circ\cdots\circ A_d)}\right) \label{eq7}
\end{equation}
where $(A_1)_{ij}=\kappa_1(\mathbf{x}_1^i,\mathbf{x}_1^j)$, $(A_2)_{ij}=\kappa_2(\mathbf{x}_2^i,\mathbf{x}_2^j)$, $\cdots$, $(A_d)_{ij}=\kappa_d(\mathbf{x}_d^i,\mathbf{x}_d^j)$, and $\circ$ denotes the Hadamard product.
\end{definition}

We use RBF kernel as recommended by~\cite{yu2018understanding}, and estimate $\mathbf{I}(\mathbf{x};\mathbf{z})$ and $\mathbf{T}(\mathbf{z})$ in each training iteration. 
Given input data in current mini-batch (of size $N$) $X=\{\mathbf{x}^1,\mathbf{x}^2,\cdots,\mathbf{x}^N\}$, suppose the corresponding latent representation is $Z=\{\mathbf{z}^1,\mathbf{z}^2,\cdots,\mathbf{z}^N\}$, in which $\mathbf{z}^i = [z^i_1, z^i_2, \cdots, z^i_d]^T$. Both $\mathbf{I}(\mathbf{x};\mathbf{z})=\mathbf{H}(\mathbf{x}) + \mathbf{H}(\mathbf{z}) - \mathbf{H}(\mathbf{x},\mathbf{z})$ and $\mathbf{T}(\mathbf{z})=\sum_{i=1}^d \mathbf{H}(z_i) - \mathbf{H}(\mathbf{z})$ can be simply evaluated based on \emph{Definition~1} and \emph{Definition~2}.

For example, in order to estimate $\mathbf{H}(\mathbf{z})$, one just needs to compute a $N\times N$ Gram matrix $A$ in which $A(i,j)$ measures the distance between latent representation $\mathbf{z}_i$ (of the $i$-th mini-batch sample) and $\mathbf{z}_j$ (of the $j$-th mini-batch sample) in the kernel space. 
Since the new estimator involves eigenvalue decomposition to matrices of $N\times N$, the scalability depends on the mini-batch size. 
Luckily, state-of-the-art VAEs are always trained with $N$ equal to $64$.

Finally, in order to obtain a smoother visualization on the information plane, we sample $6,400$ data points and obtain the latent representation by feeding them into the Gaussian type encoder. We then use the matrix-based R{\'e}nyi's $\alpha$-entropy functional to compute $\mathbf{I}(\mathbf{x};\mathbf{z})$ and $\mathbf{T}(\mathbf{z})$ in a mini-batch of size $64$ to obtain $100$ estimation values to both quantities. We finally average the $100$ estimation values to obtain a stable result.

\begin{table}[h]
\caption{The unified network architecture in Section 6.1 experiments.} \label{tb:architecture}
\renewcommand{\arraystretch}{1.2}
\begin{center}
\scalebox{1.0}{
\begin{tabular}{ll}
\hline
\textbf{Encoder} & \textbf{Decoder} \\ \hline
FC, $2 \times 10$ (mean, log variance) & Input: $\mathbb{R}^{10}$ \\
FC, 256 ReLU & FC, 256 ReLU \\
$4 \times 4$ Conv, stride 2, 64 ReLU & FC, $4 \times 4 \times 64$ ReLU \\
$4 \times 4$ Conv, stride 2, 64 ReLU & $4 \times 4 $ Upconv, stride 2, 64 ReLU \\
$4 \times 4$ Conv, stride 2, 32 ReLU & $4 \times 4 $ Upconv, stride 2, 32 ReLU \\
$4 \times 4$ Conv, stride 2, 32 ReLU & $4 \times 4 $ Upconv, stride 2, 32 ReLU \\
Input: $64 \times 64 \times$ \#channel & $4 \times 4 $ Upconv, stride 2, \#channel \\ \hline
\end{tabular}}
\end{center}
\end{table}

\begin{table}[h]
\caption{The network architecture for the discriminator in FactorVAE.}\label{tb:discriminator}
\centering
\scalebox{1.1}{
\begin{tabular}{l}
\hline
Discriminator \\ \hline
FC, 1000 leaky ReLU \\
FC, 1000 leaky ReLU \\
FC, 1000 leaky ReLU \\
FC, 1000 leaky ReLU \\
FC, 1000 leaky ReLU \\
FC, 1000 leaky ReLU \\
FC, 2 \\ \hline
\end{tabular}}
\end{table}

\subsection{Quantitative Evaluation}
In the quantitative comparison experiments, we use a unified network architecture, which is shown in the Table~\ref{tb:architecture}.
A Gaussian type encoder outputs the mean and the log variance given the image data, and a Bernoulli type decoder takes the mean as the input and outputs the generated images.
The models are optimized by the Adam optimiser with learning rate \num{1e-4}, $\beta_{1}=0.9$, $\beta_{2}=0.999$, the batch size is fixed as \num{64} and the total training steps are \num{300000} for both of \textit{dSprites} and \textit{Cars3D} data sets. Table~\ref{tb:discriminator} lists the discriminator architecture for FactorVAE, and we use the Adam optimiser with learning rate \num{1e-4}, $\beta_{1}=0.5$, $\beta_{2}=0.9$, $\epsilon=\num{1e-8}$ for optimization.

\subsection{Qualitative Evaluation}
Table~\ref{tb:Qualitative_architecture} shows the network architecture used in the Section 6.2 for qualitative evaluation on \textit{3D Chairs} and \textit{Fashion-MNIST} data sets.
The following sections list the training details respectively.

\begin{table}[h]
\caption{The network architecture in Section 6.2 experiments.} \label{tb:Qualitative_architecture}
\vspace{-0.5em}
\renewcommand{\arraystretch}{1.2}
\begin{center}
\scalebox{1.0}{
\begin{tabular}{ll}
\hline
\textbf{Encoder} & \textbf{Decoder} \\ \hline
FC, $2 \times 10$ (mean, log variance) & Input: $\mathbb{R}^{10}$ \\
2FCs, 256 ReLU & 2FCs, 256 ReLU \\
$4 \times 4$ Conv, stride 2, 32 ReLU & FC, $4 \times 4 \times 32$ ReLU \\
$4 \times 4$ Conv, stride 2, 32 ReLU & $4 \times 4 $ Upconv, stride 2, 32 ReLU \\
$4 \times 4$ Conv, stride 2, 32 ReLU & $4 \times 4 $ Upconv, stride 2, 32 ReLU \\
$4 \times 4$ Conv, stride 2, 32 ReLU & $4 \times 4 $ Upconv, stride 2, 32 ReLU \\
Input: $64 \times 64 \times$ \#channel & $4 \times 4 $ Upconv, stride 2, \#channel \\ \hline
\end{tabular}}
\end{center}
\vspace{-0.5em}
\end{table}

\subsubsection{3D Chairs}
\begin{itemize}
    \item Decoder type: Bernoulli
    \item Batch size: 64
    \item Training epochs: 300
    \item Optimizer: Adam with learning rate \num{1e-4}
\end{itemize}

\subsubsection{Fashion-MNIST}
\begin{itemize}
    \item Decoder type: Bernoulli
    \item Batch size: 64
    \item Training epochs: 400
    \item Optimizer: Adam with learning rate \num{1e-4}
\end{itemize}


\section{Additional Results}
\subsection{Empirical behavior of $\mathbf{I}(\mathbf{x};\mathbf{z})$ and $\mathbf{T}(\mathbf{z})$}
\begin{figure}[h]
    \centering
    \subfloat{\includegraphics[width=0.3\textwidth]{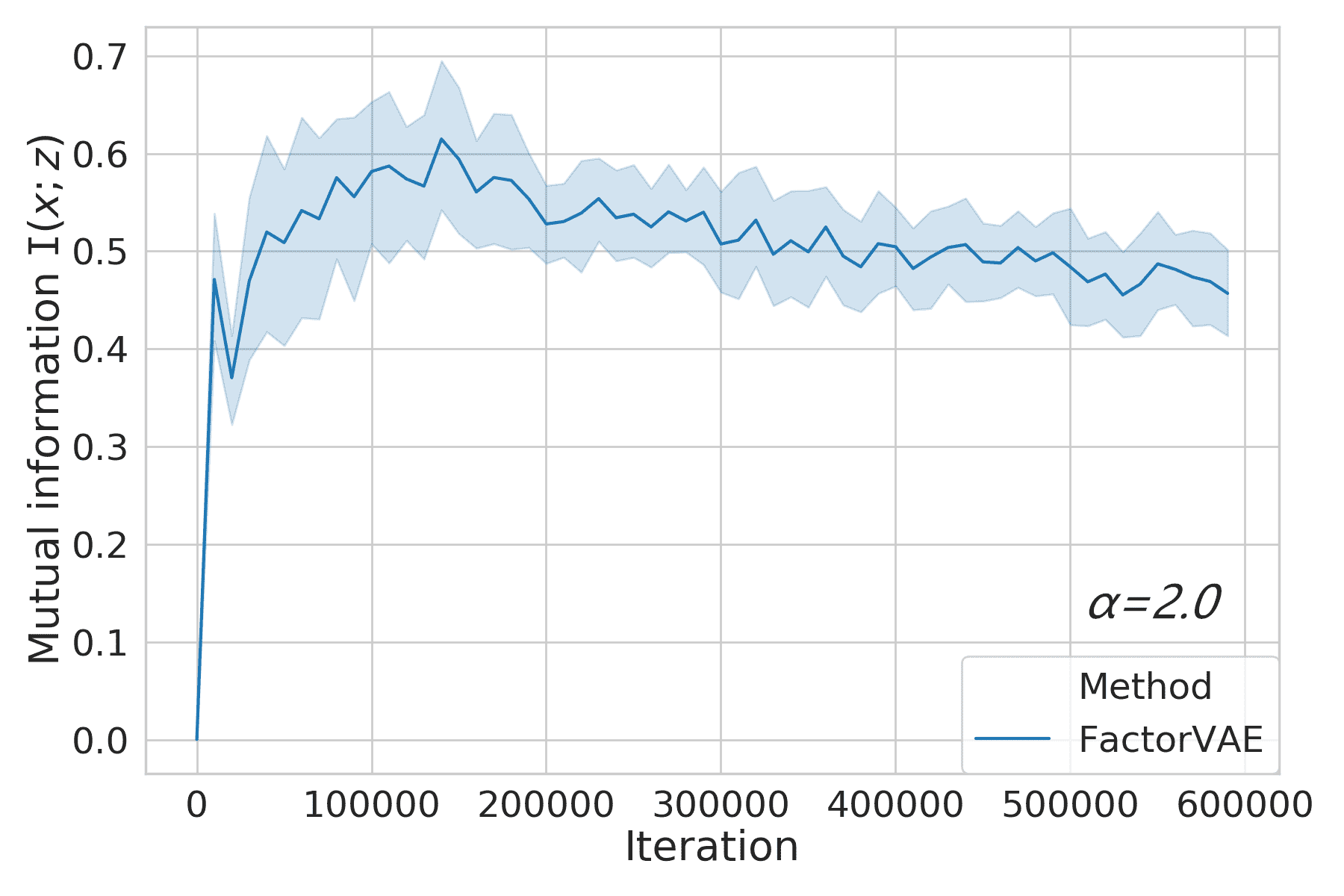}} \hspace{0.02\textwidth}
    \subfloat{\includegraphics[width=0.3\textwidth]{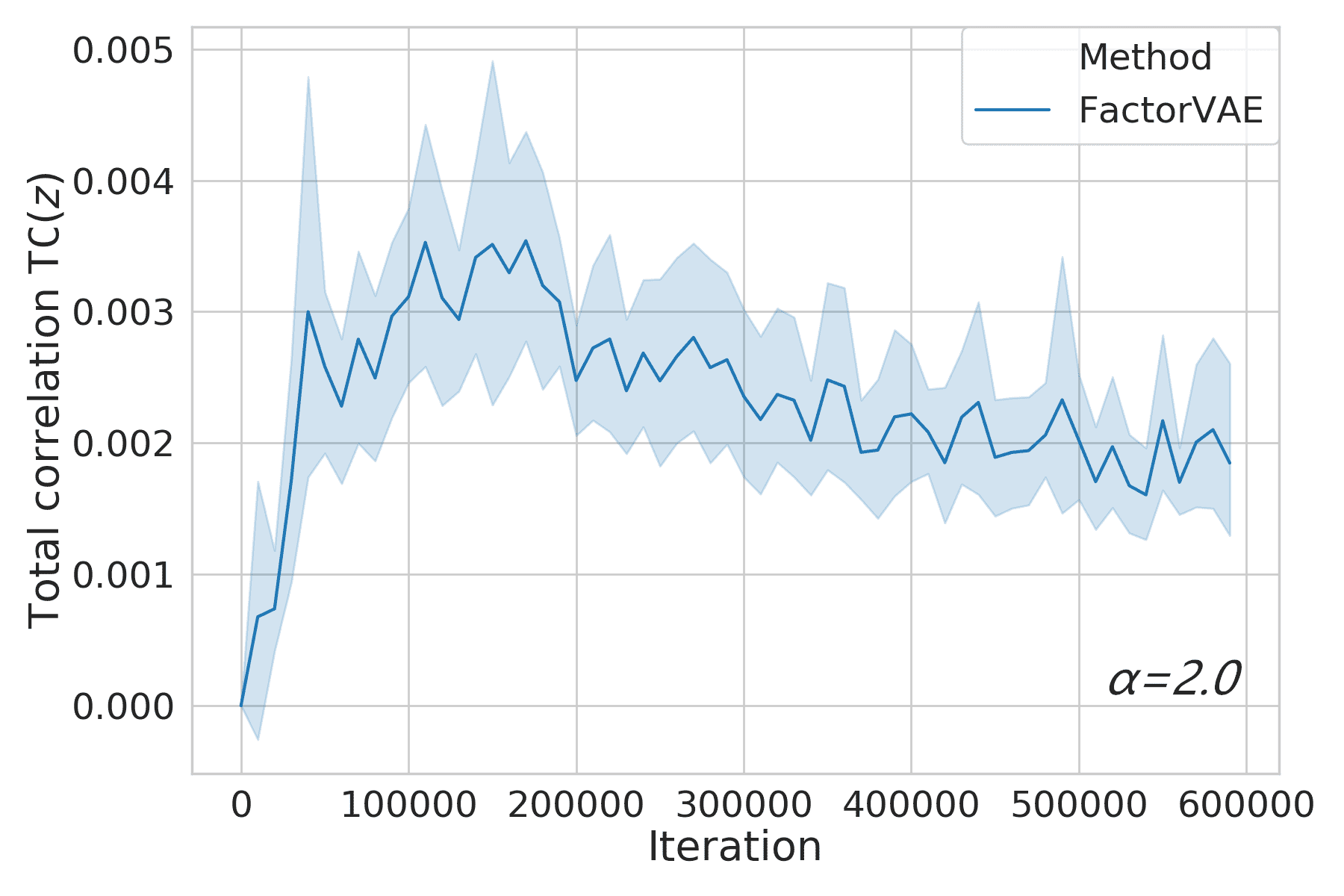}} \hspace{0.02\textwidth}
    \subfloat{\includegraphics[width=0.24\textwidth]{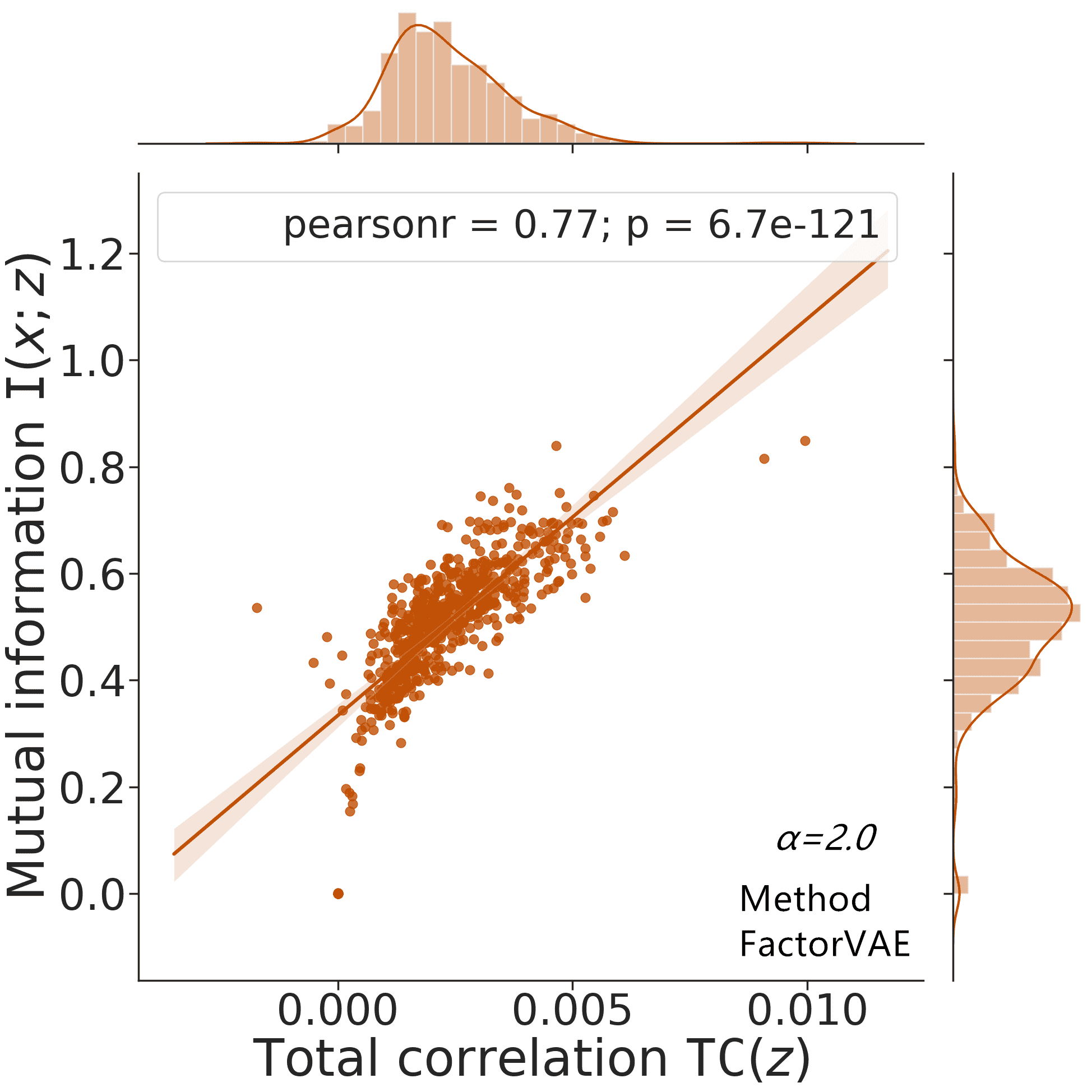}}\\
    \subfloat{\includegraphics[width=0.3\textwidth]{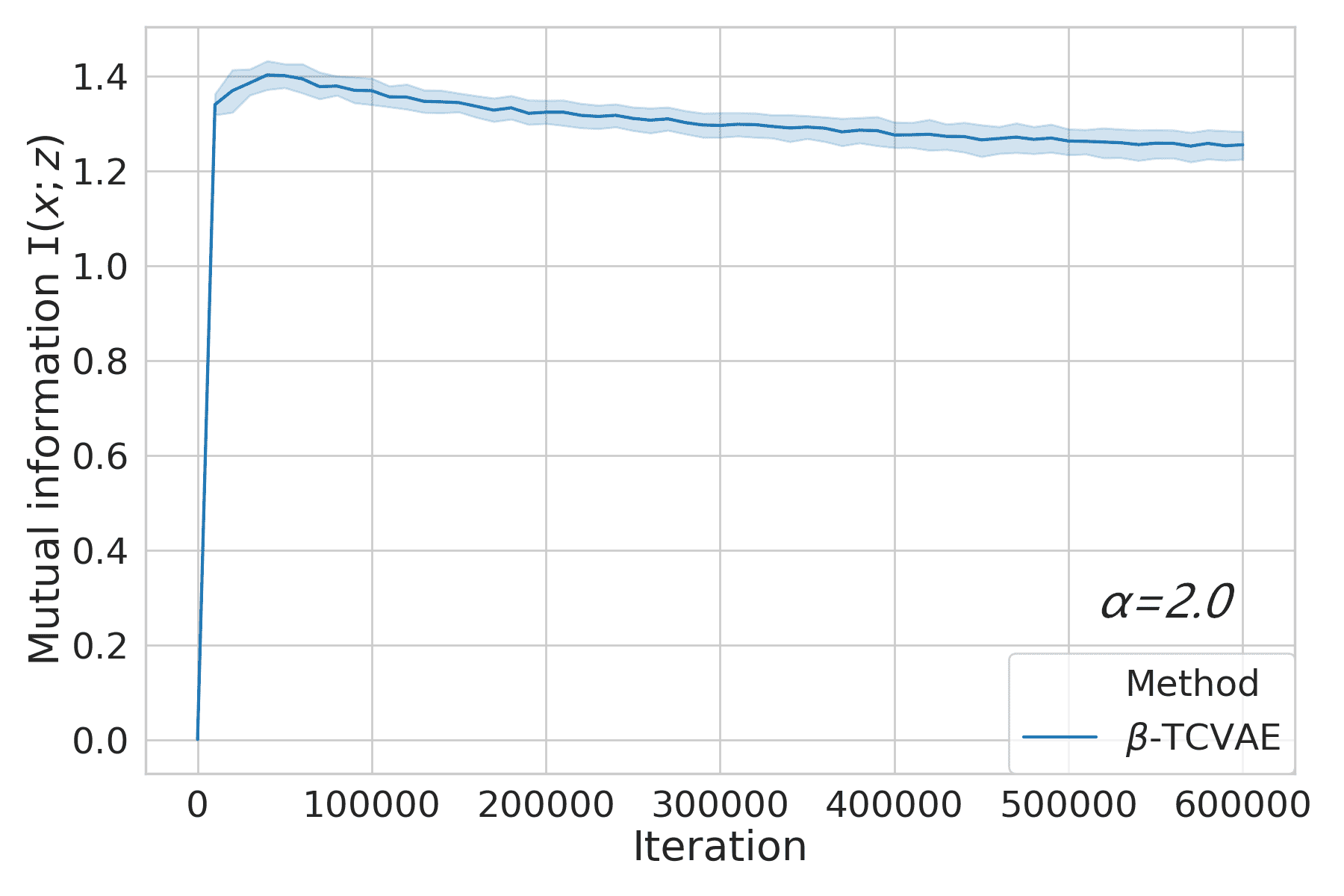}} \hspace{0.02\textwidth}
    \subfloat{\includegraphics[width=0.3\textwidth]{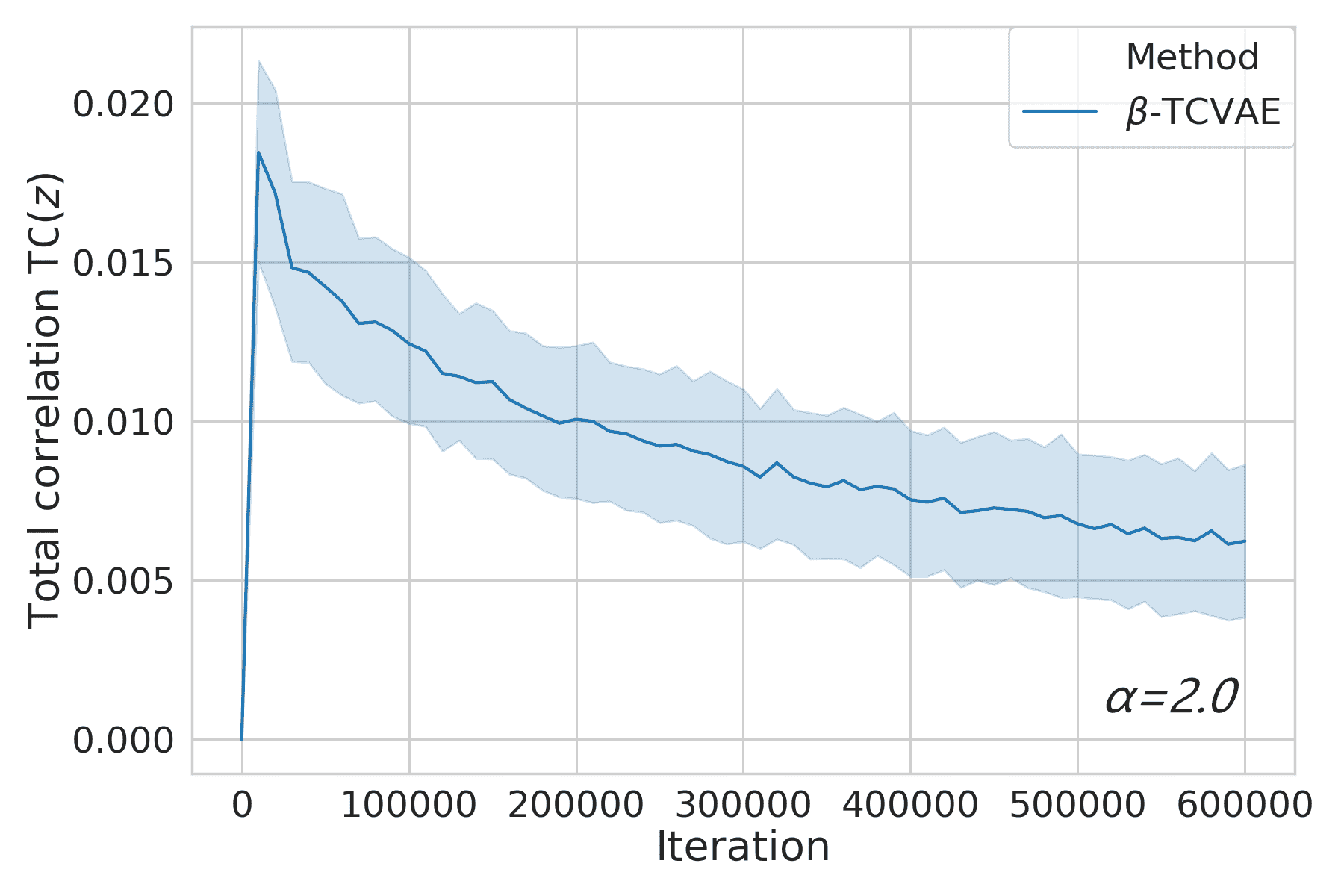}} \hspace{0.02\textwidth}
    \subfloat{\includegraphics[width=0.24\textwidth]{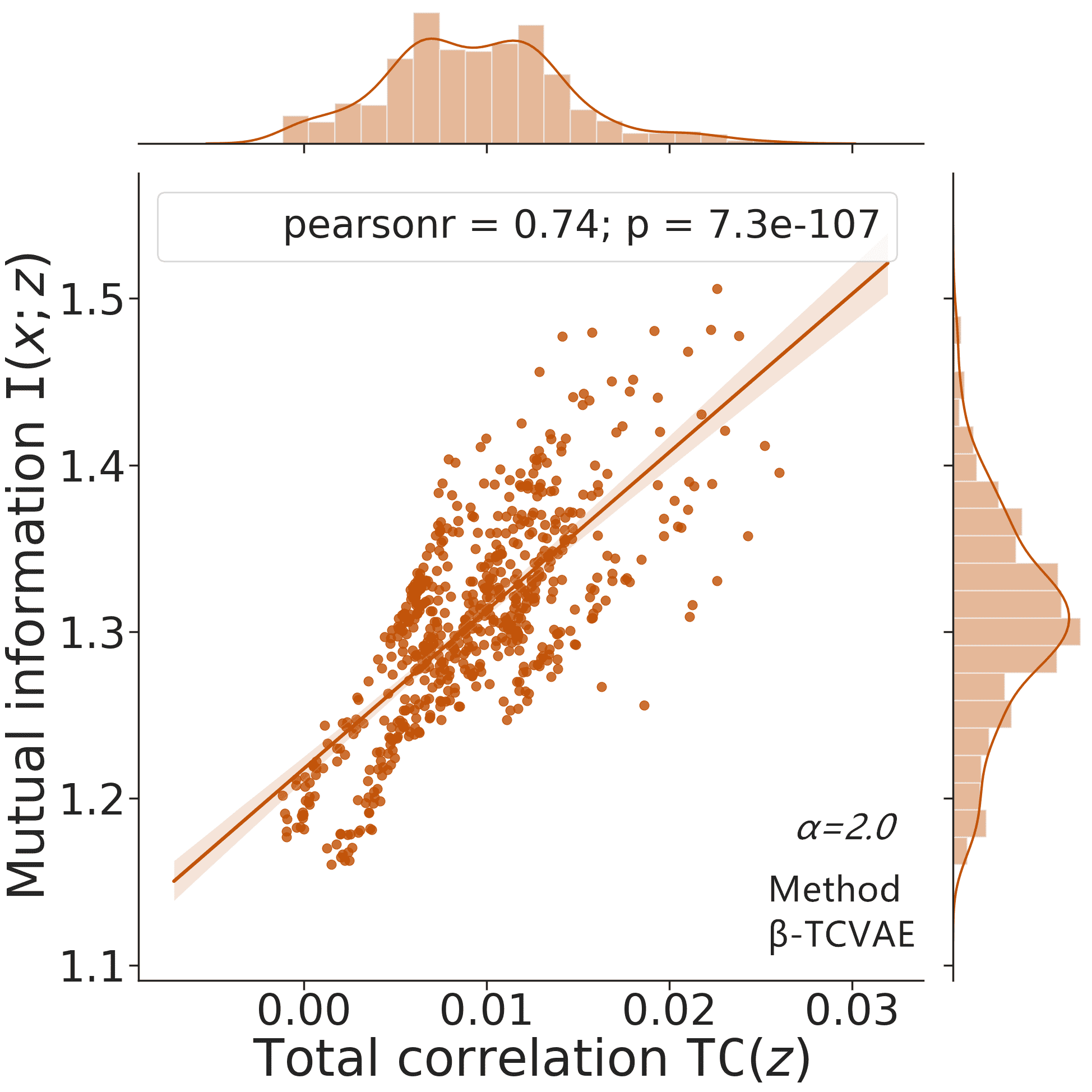}}\\
    \subfloat{\includegraphics[width=0.3\textwidth]{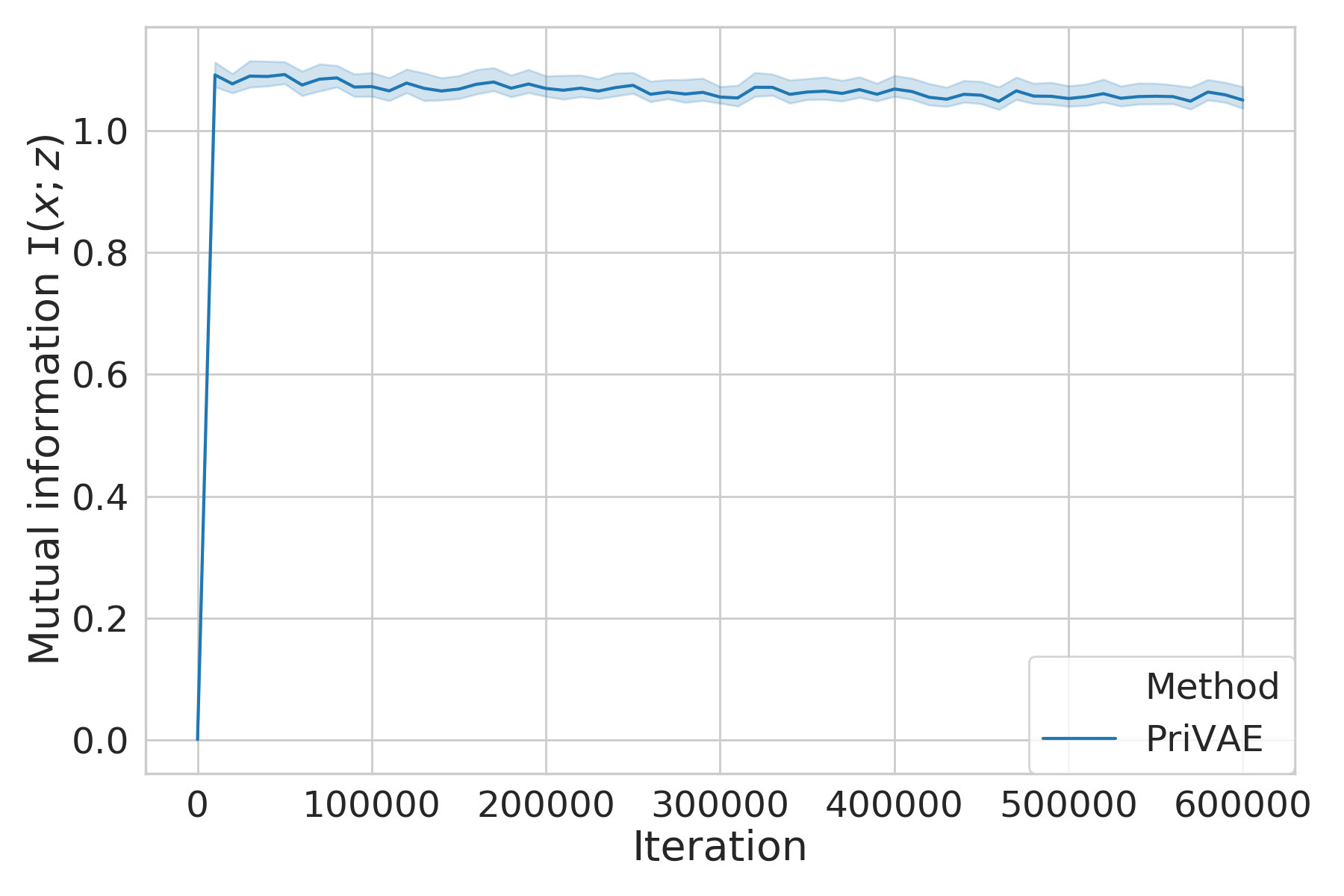}} \hspace{0.02\textwidth}
    \subfloat{\includegraphics[width=0.3\textwidth]{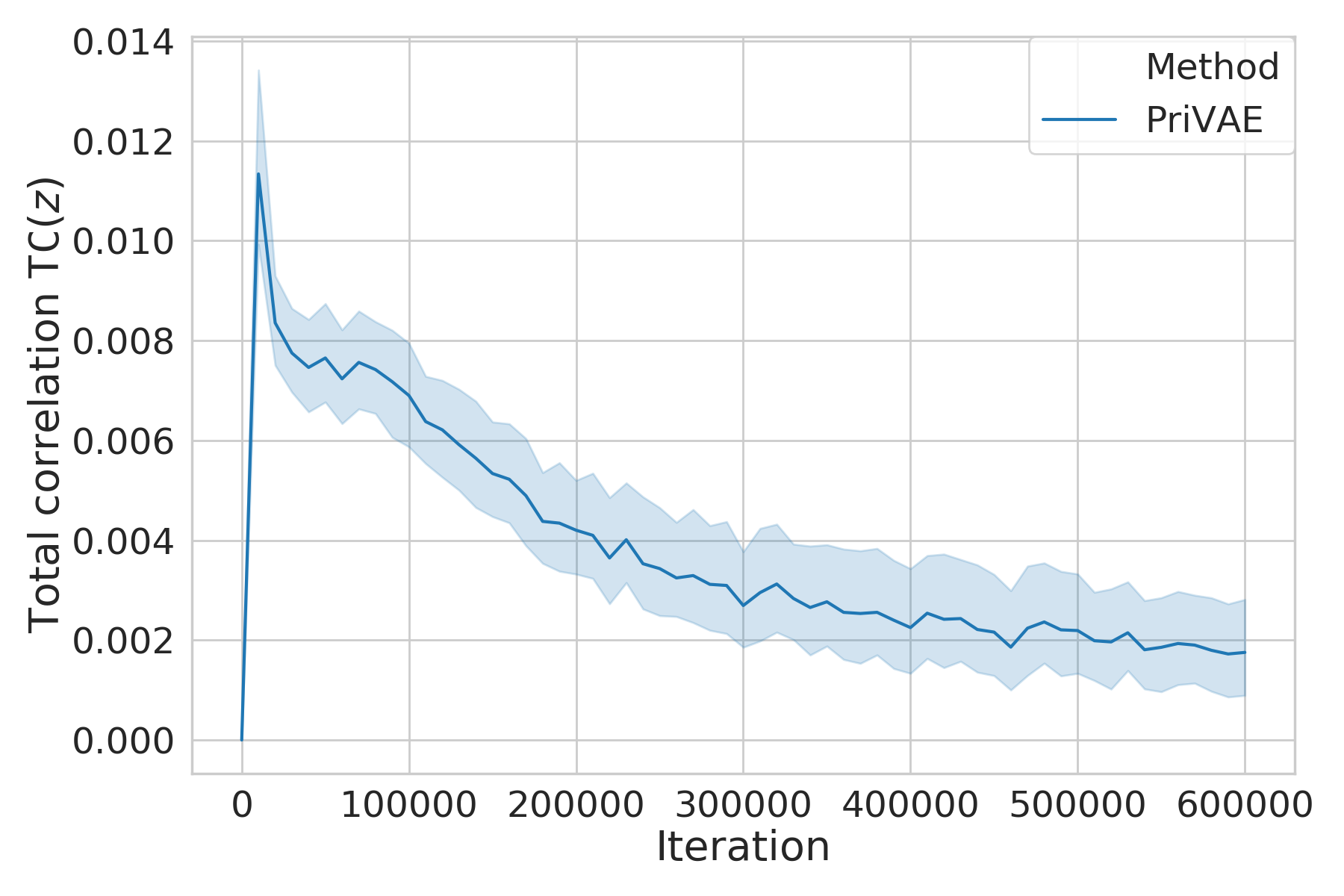}} \hspace{0.02\textwidth}
    \subfloat{\includegraphics[width=0.24\textwidth]{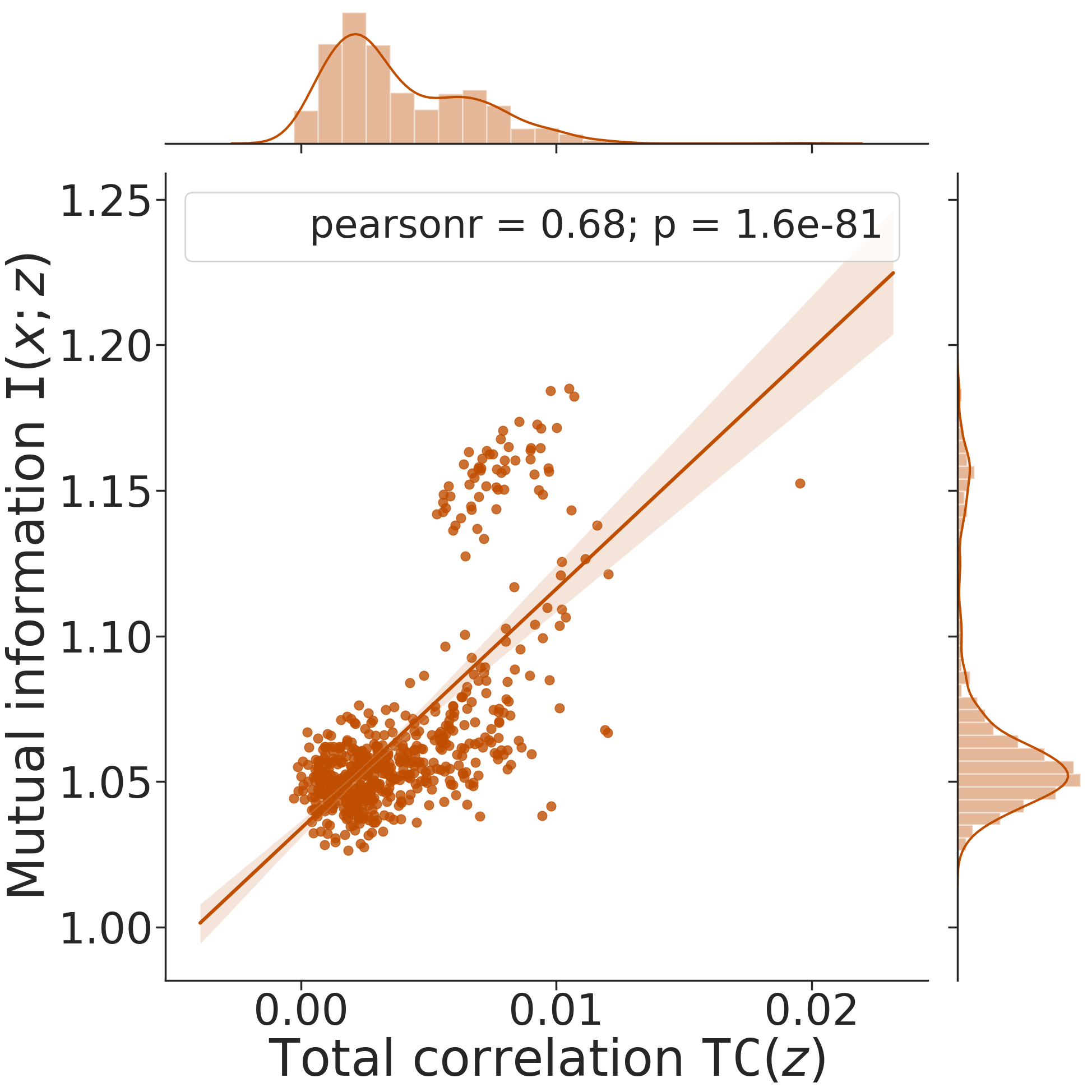}}\\
    
    \caption{Measured by the matrix-based R{\'e}nyi's $\alpha$-entropy estimator with $\alpha=2.0$, the evolution of the value of $\mathbf{I}(\mathbf{x};\mathbf{z})$ (left column) and the value of $\mathbf{T}(\mathbf{z})$ (middle column) across training iterations for FactorVAE (first row), $\beta$-TCVAE (second row) and PRI-VAE (third row) on \textit{Cars3d} data set. The right column shows the positive correlation between these two values.} 
    \label{fig:learning_curve_of_MI_TC_alpha-2}
\end{figure}

\begin{figure}[h]
    \centering
    \subfloat{\includegraphics[width=0.3\textwidth]{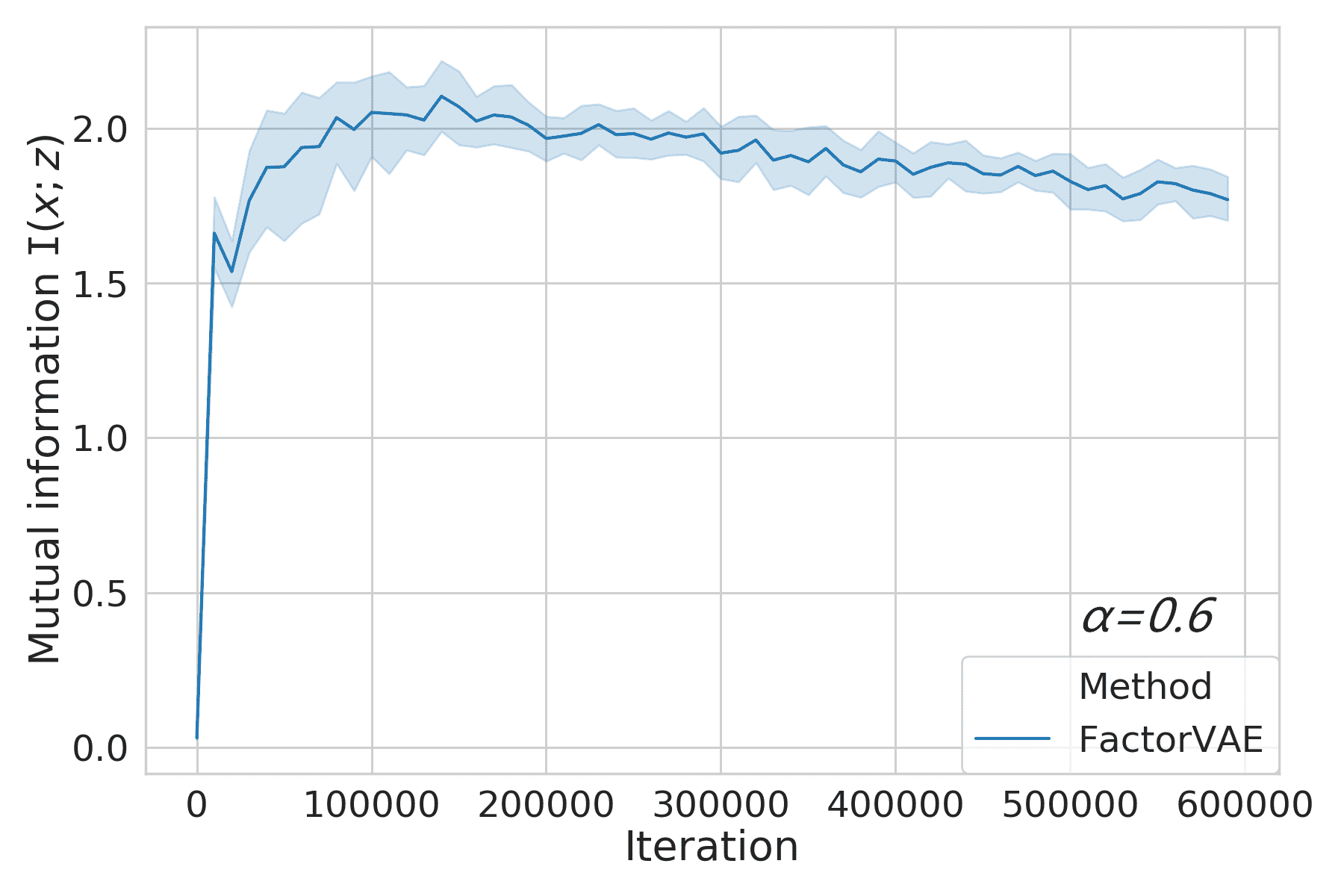}} \hspace{0.02\textwidth}
    \subfloat{\includegraphics[width=0.3\textwidth]{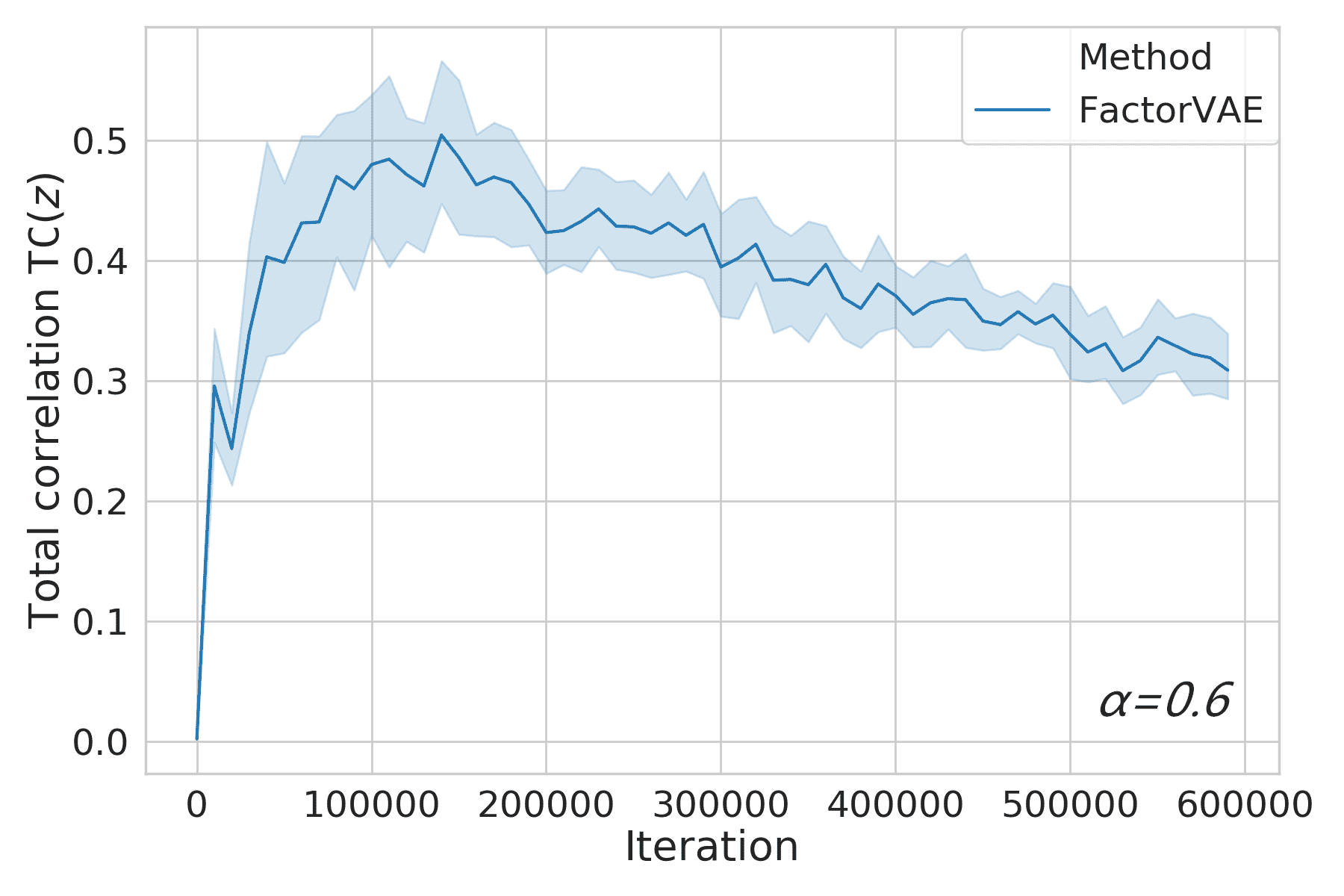}} \hspace{0.02\textwidth}
    \subfloat{\includegraphics[width=0.24\textwidth]{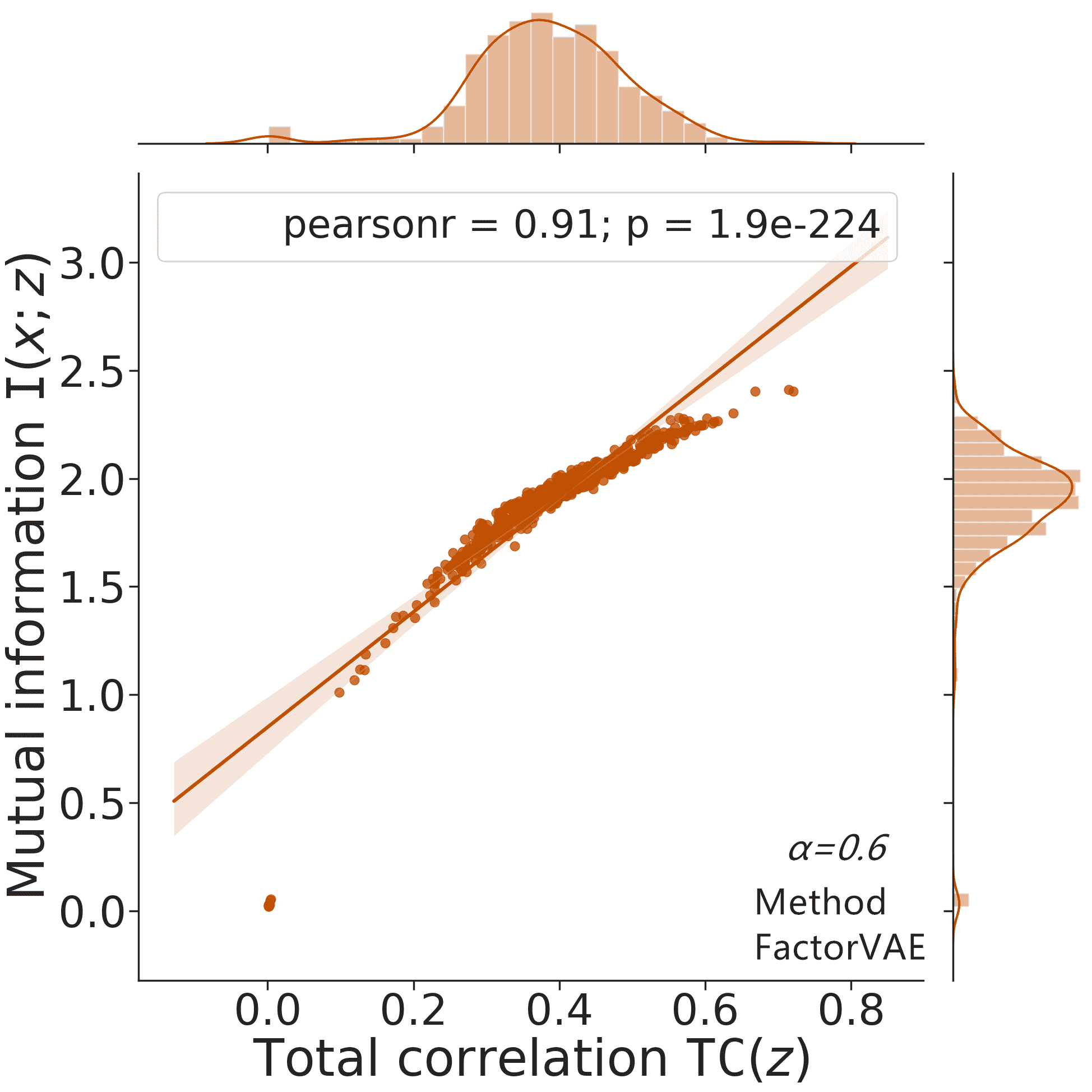}}\\
    \subfloat{\includegraphics[width=0.3\textwidth]{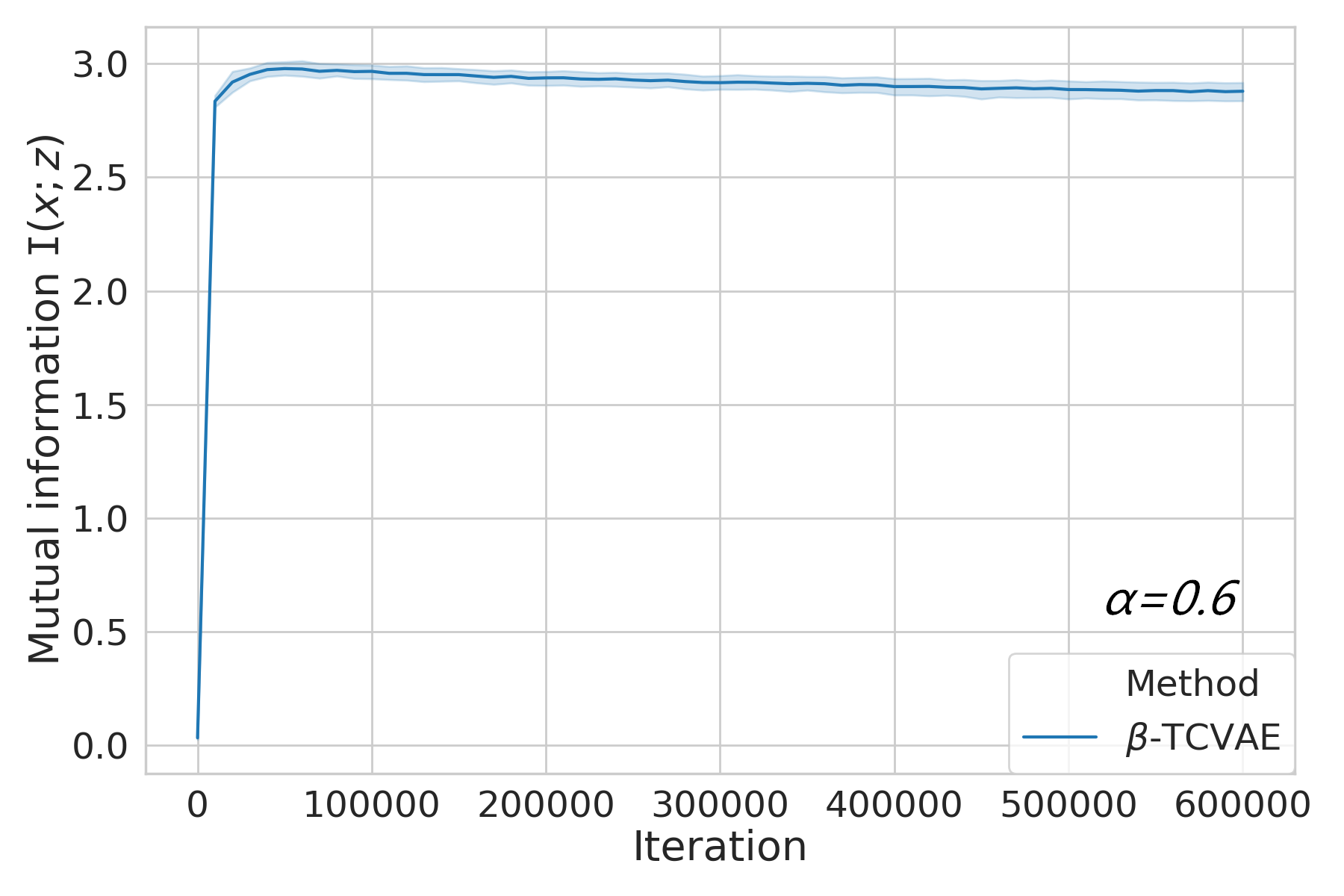}} \hspace{0.02\textwidth}
    \subfloat{\includegraphics[width=0.3\textwidth]{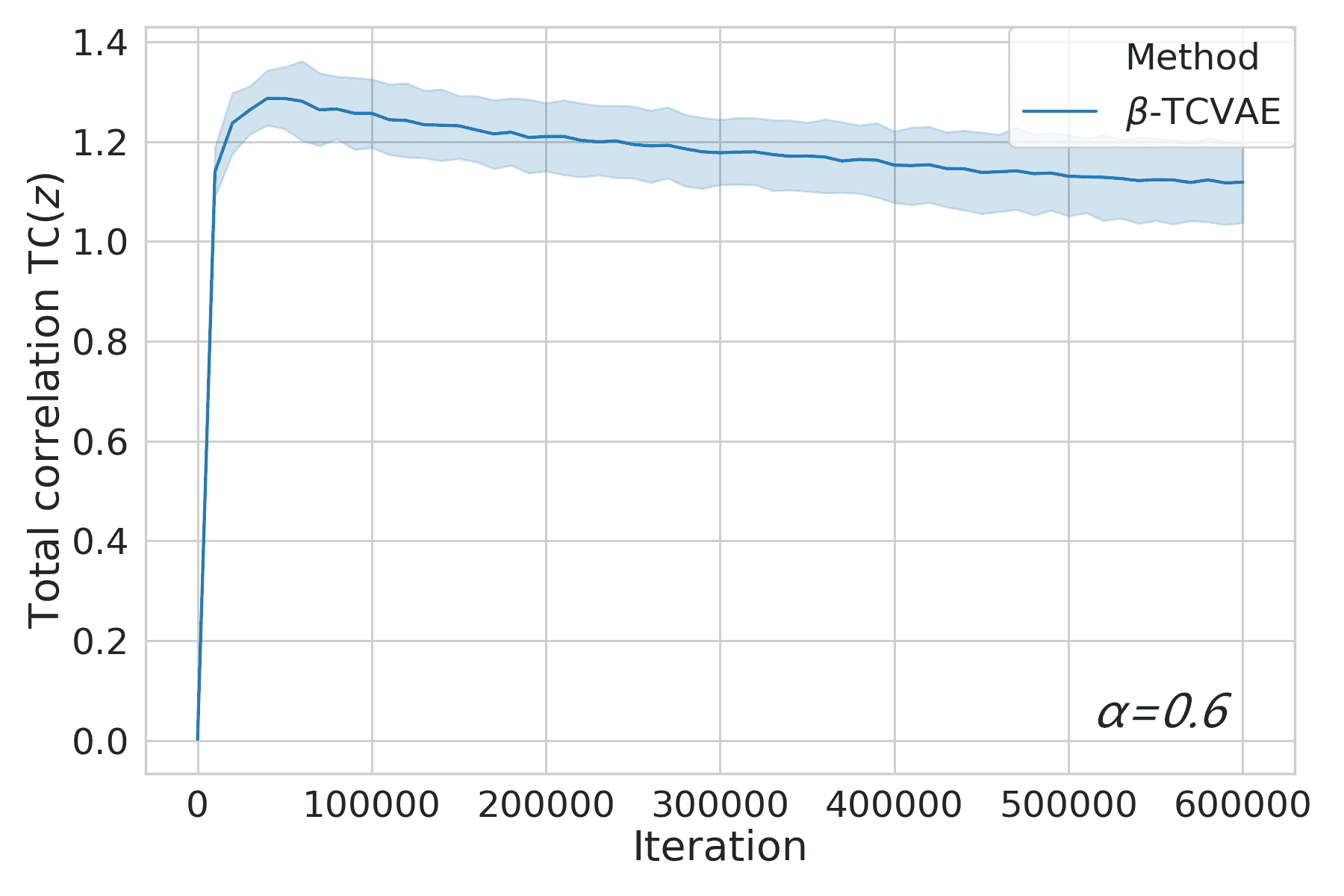}} \hspace{0.02\textwidth}
    \subfloat{\includegraphics[width=0.24\textwidth]{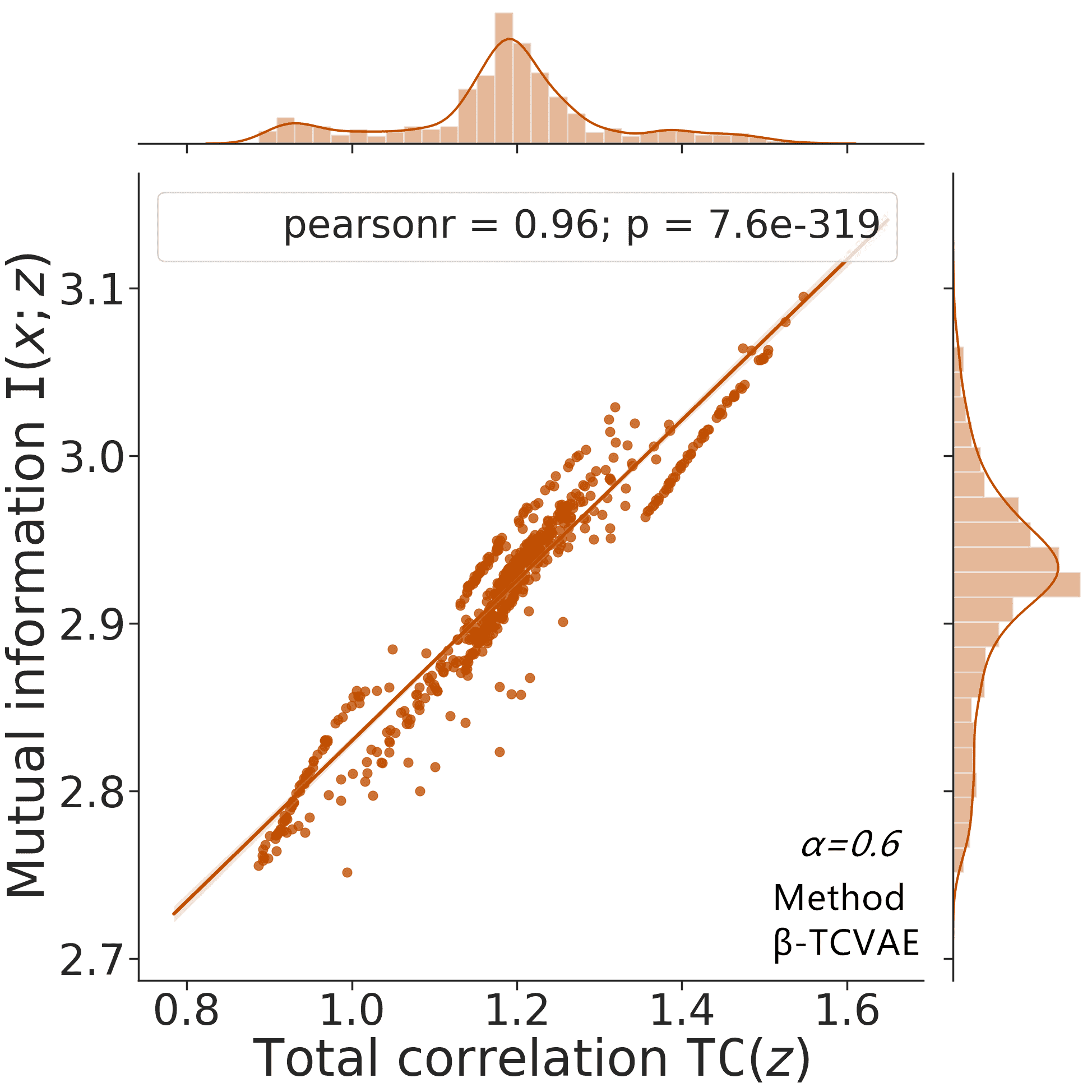}}\\
    \subfloat{\includegraphics[width=0.3\textwidth]{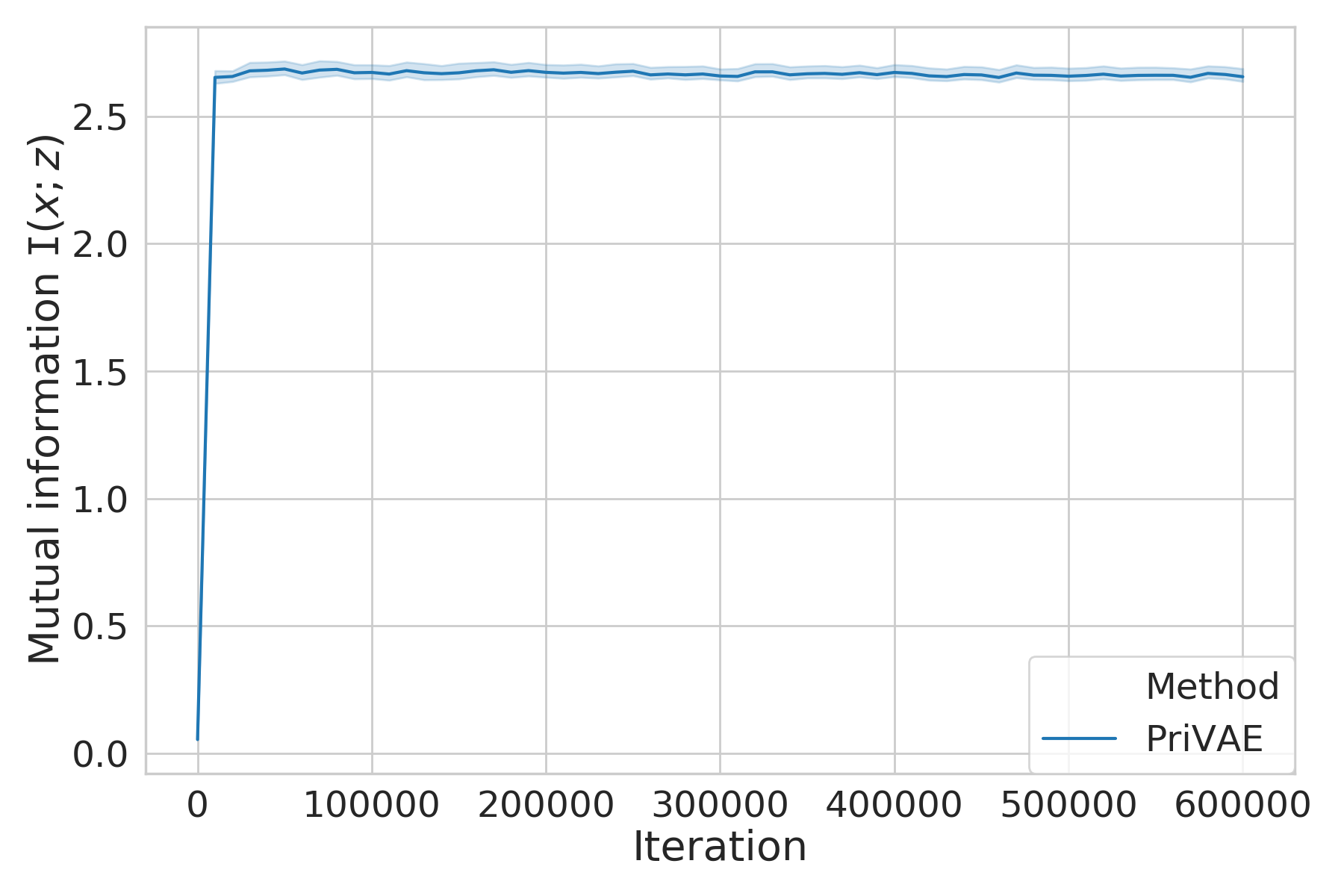}} \hspace{0.02\textwidth}
    \subfloat{\includegraphics[width=0.3\textwidth]{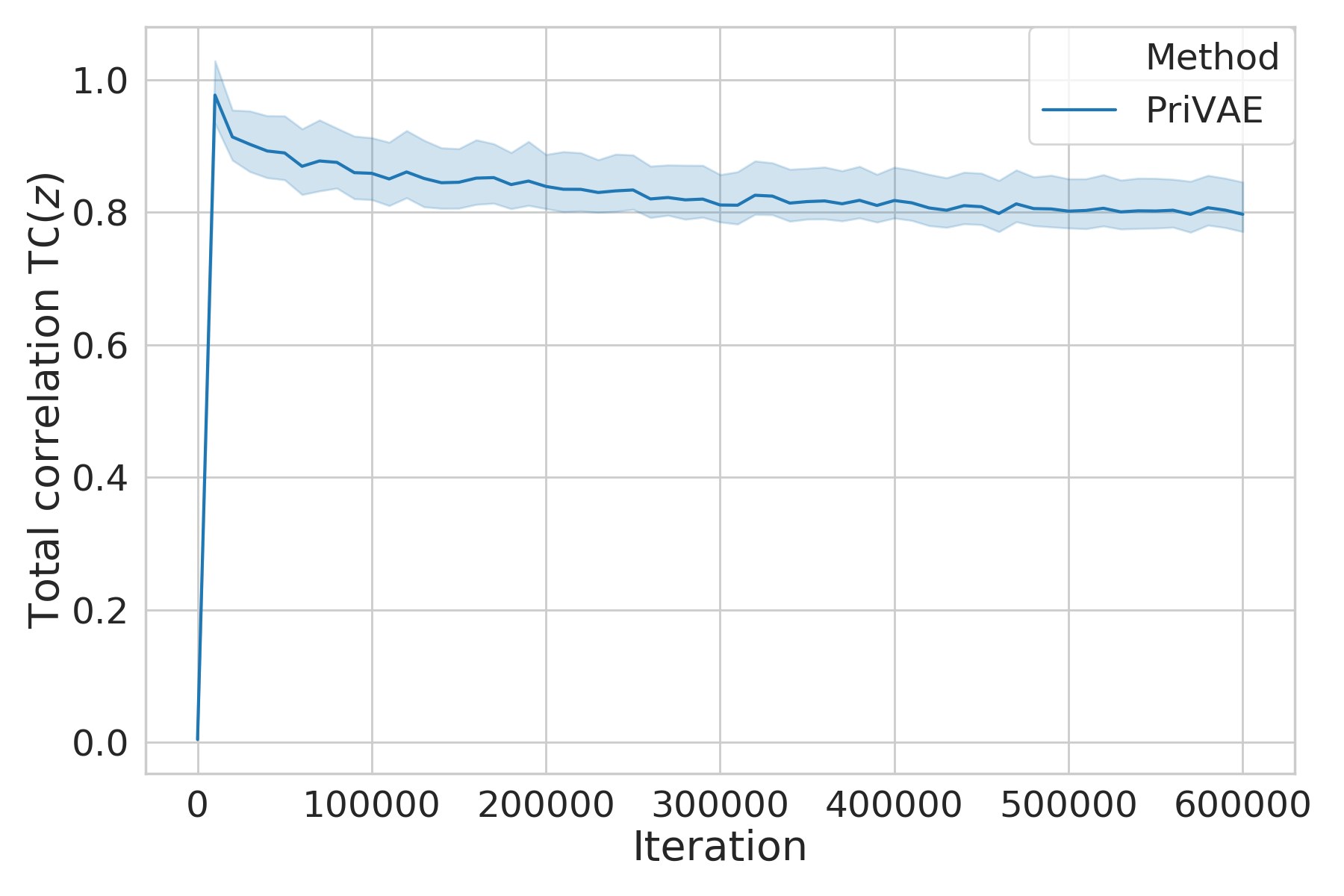}} \hspace{0.02\textwidth}
    \subfloat{\includegraphics[width=0.24\textwidth]{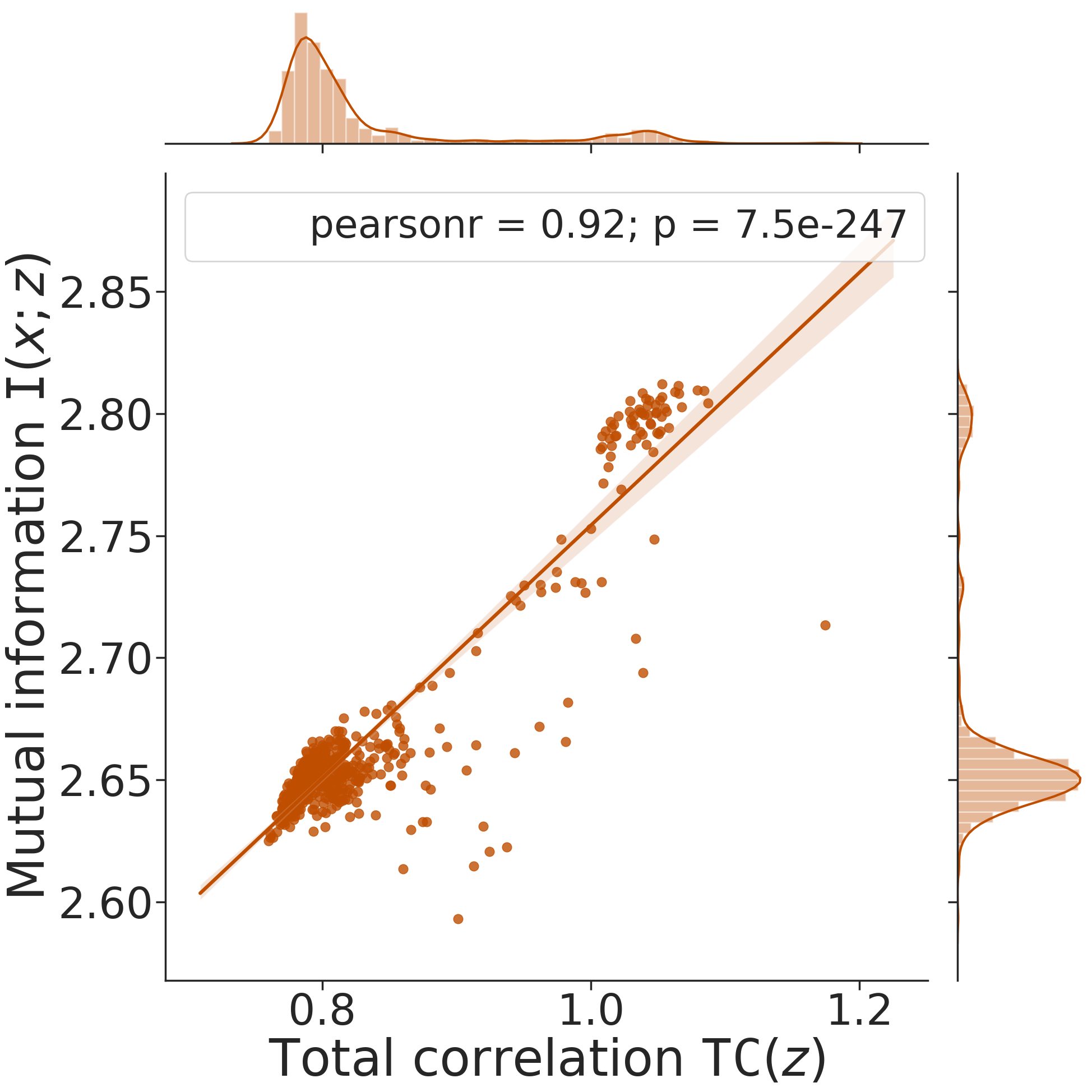}}\\
    
    \caption{Measured by the matrix-based R{\'e}nyi's $\alpha$-entropy estimator with $\alpha=0.6$, the evolution of the value of $\mathbf{I}(\mathbf{x};\mathbf{z})$ (left column) and the value of $\mathbf{T}(\mathbf{z})$ (middle column) across training iterations for FactorVAE and $\beta$-TCVAE on \textit{Cars3d} data set. The right column shows the positive correlation between these two values.} 
    \label{fig:learning_curve_of_MI_TC_alpha-0-6}
\end{figure}

The matrix-based R{\'e}nyi's $\alpha$-entropy introduces a hyper-parameter $\alpha$ when estimating the entropy value of the input.
To verify the consistency of the observed general trend of dynamic of learning presented in Section~\ref{sec:Empirical_behavior} and reduce the possible effects introduces by different hyper-parameter setting, we demonstrate the empirical behavior of $\mathbf{I}(\mathbf{x};\mathbf{z})$ and $\mathbf{T}(\mathbf{z})$ with additional $\alpha$ setting for matrix-based R{\'e}nyi's $\alpha$-entropy estimator.
Fig.~\ref{fig:learning_curve_of_MI_TC_alpha-2} and Fig.~\ref{fig:learning_curve_of_MI_TC_alpha-0-6} illustrate observation results with $\alpha=2.0$ and $\alpha=0.6$ respectively. 
We can find that our general observation is accordance with different hyper-parameter settings. 

\subsection{DCI disentanglement metric}
\begin{figure}[h]
    \centering
    \subfloat{\includegraphics[width=0.45\textwidth]{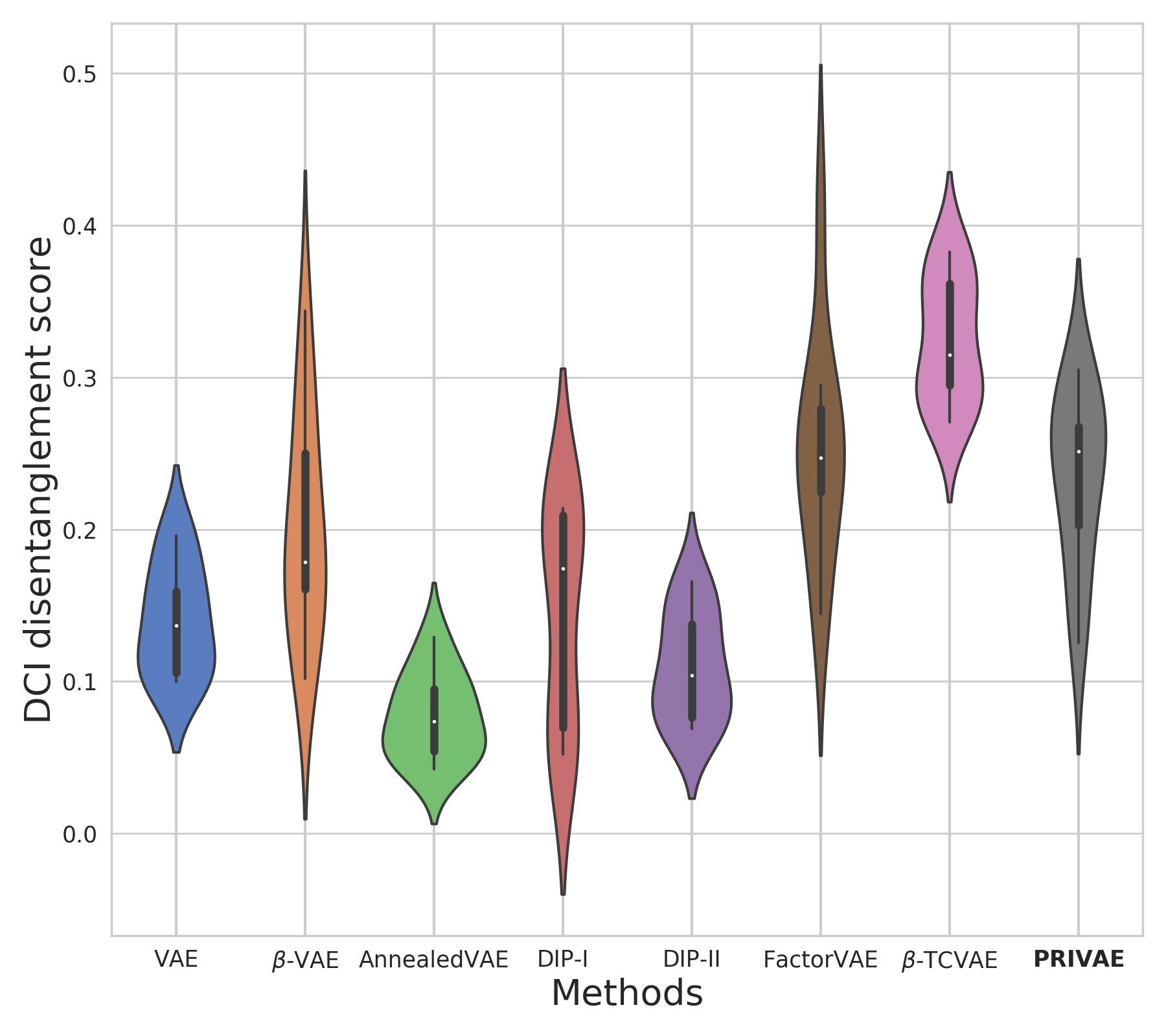}} \hspace{0.03\textwidth}
    \subfloat{\includegraphics[width=0.45\textwidth]{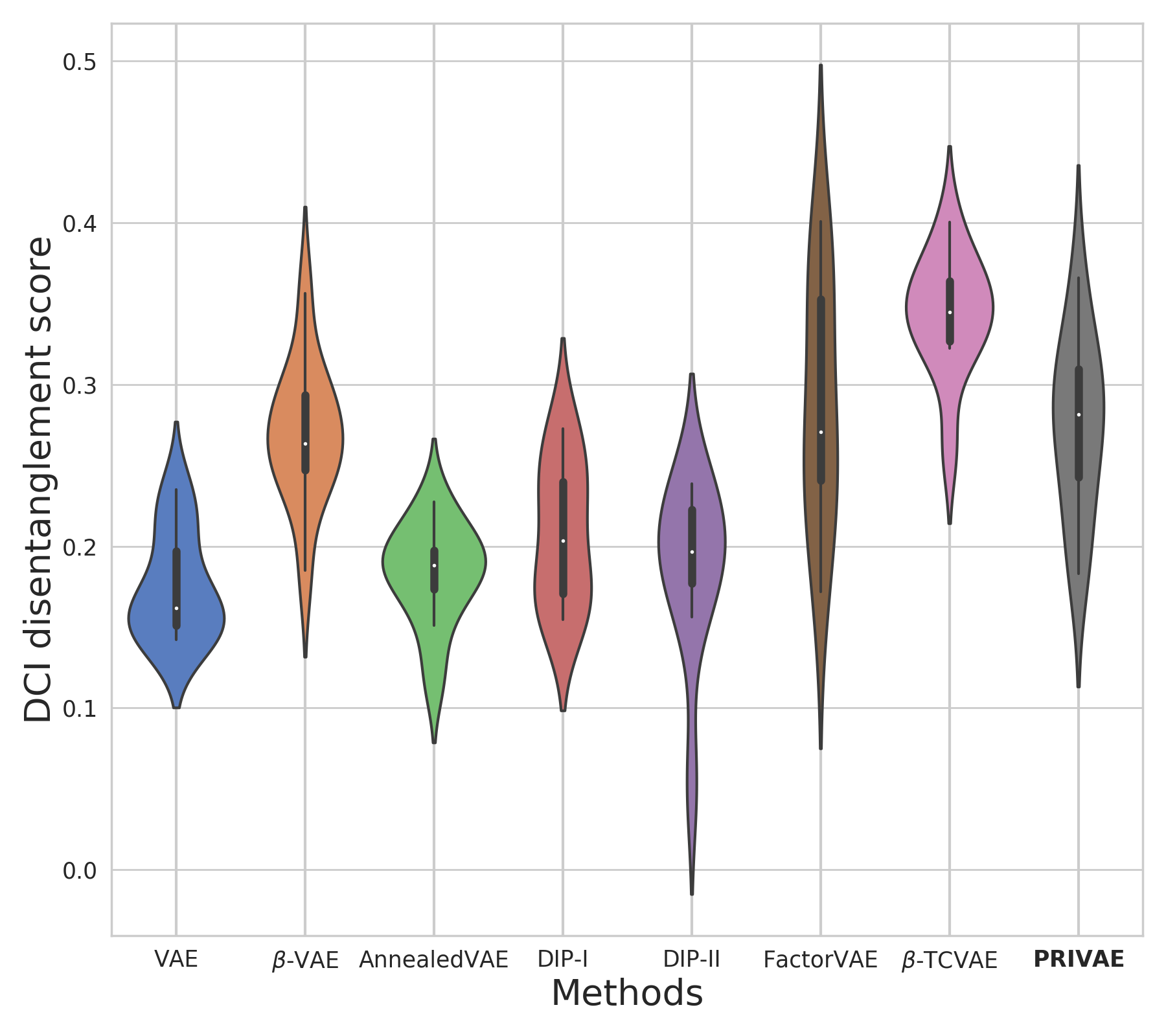}} 
    \caption{The DCI disentanglement metric on \textit{dSprites} (left) and \textit{Cars3D} (right) data sets.} 
    \label{fig:compare_baseline_dci}
\end{figure}

Fig.~\ref{fig:compare_baseline_dci} demonstrates the DCI disentanglement metric score for PRI-VAE and all other competing models in \textit{dSprites} and \textit{Cars3D}.
It shows the similar comparison results as illustrated and discussed in Section~\ref{sec:PRI_synthetic_comparison2}, where PRI-VAE achieves consistent better disentanglement performance than VAE, $\beta$-VAE, AnnealedVAE, and DPI-VAE, but worse performance than FactorVAE and $\beta$-TCVAE.

Table~\ref{tb:pri_vae_star_dci} shows the DCI disentanglement scores of $\beta$-TCVAE, PRI-VAE and PRI-VAE$^\star$ on \textit{dSprites} and \textit{Cars3D}. 
Similar comparison results are obtained and discussed in Section~\ref{sec:PRI_synthetic_comparison2}.
\begin{table}[h]
\caption{The DCI disentanglement scores of $\beta$-TCVAE, PRI-VAE and PRI-VAE$^\star$ on \textit{dSprites} and \textit{Cars3D}. We report the mean value over $10$ runs and the standard deviation in parentheses.} \label{tb:pri_vae_star_dci}
\renewcommand{\arraystretch}{1.5}
\begin{center}
\begin{tabular}{llll}
\hline
 & $\beta$-TCVAE & PRI-VAE & PRI-VAE$^\star$ \\ \hline
\textit{dSprites} & 0.318 (0.040) & 0.234 (0.058) & 0.345 (0.088) \\ \hline
\textit{Cars3D} & 0.343 (0.037) & 0.281 (0.050) & 0.321 (0.048) \\ \hline
\end{tabular}
\end{center}
\end{table}

\subsection{Hyper-parameter Analysis} \label{sec:hyperparameter}
\begin{figure}[h]
    \centering
    \includegraphics[width=0.9\textwidth]{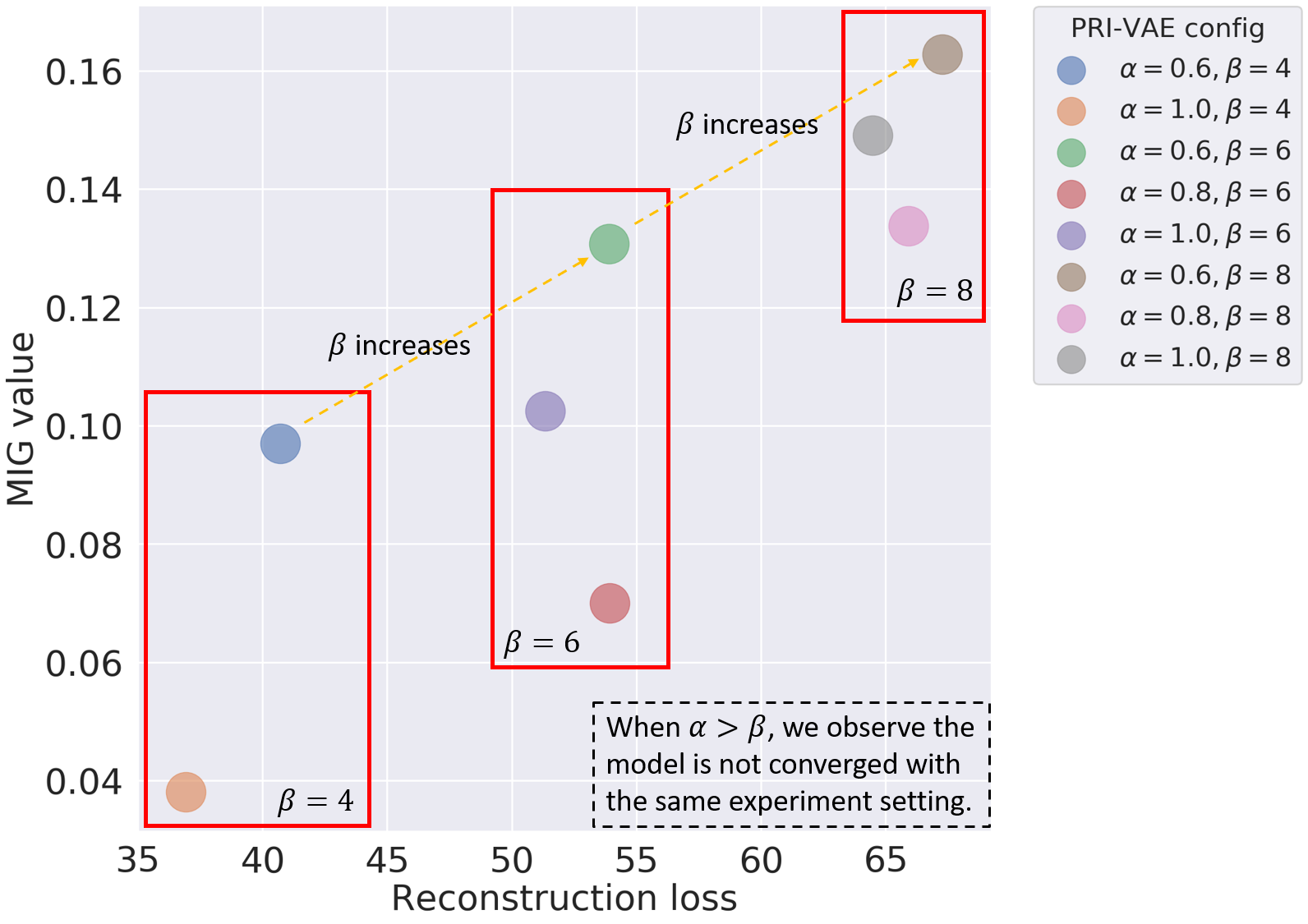}
    \caption{The disentanglement and reconstruction trade-off with respect to different parameter setting of $\alpha$ and $\beta$ in PRI-VAE. When $\alpha>\beta$, the model does not converge.}
    \label{fig:PriVAE_hyperparameters}
    \vspace{-0.5em}
\end{figure}

Fig.~\ref{fig:PriVAE_hyperparameters} shows the MIG score and the reconstruction loss measured at the end of training with respect to different values of $\beta$ and $\alpha=1$ in~\textit{dSprites} data set. Obviously, with the increase of $\beta$, the model disentanglement performance increases accordingly. This makes sense, because we penalize more on the independence between each dimension of $z$. However, we also observed an obvious trade-off before disentanglement and reconstruction, which has also been mentioned in previous works (e.g.,~\cite{kim2018disentangling,chen2018isolating}). We intend to address this problem in the future. A promising solution comes from~\cite{lezama2018overcoming}.

On the other hand, it is worth noting that our model is hard to convergence when $\alpha>\beta$. This result matches well with the property of PRI. As discussed earlier, $\beta/\alpha$ plays the same role as $\gamma$ in PRI. Theoretically, PRI with $\gamma=1$ reduces to the classical mean shift algorithm~\cite{rao2009mean} such that it will iteratively reaching to the mode of the prior distribution. Note that, the mode of an isotropic Gaussian is exactly its mean vector (a single point in latent space). Therefore, $\beta/\alpha=1$ pushes our latent representation to a single point. Obviously, this will seriously impedes our training. In fact, as demonstrated in Fig.~1 in the main text, $\beta/\alpha\geq2$ is able to balance a good trade-off between structure preservation and uncertainty filtering. Therefore, we recommend $\beta\geq2\alpha$.

In this section, we investigate the effects of hyper-parameters $\alpha$ and $\beta$ to the performance of PRI-VAE. 

\onecolumn

\subsection{More Qualitative Results}

\begin{figure}[!h]
    \centering
    \subfloat[The reconstruction results of PRI-VAE on \textit{dsprites} data set.]{\includegraphics[width=.9\textwidth]{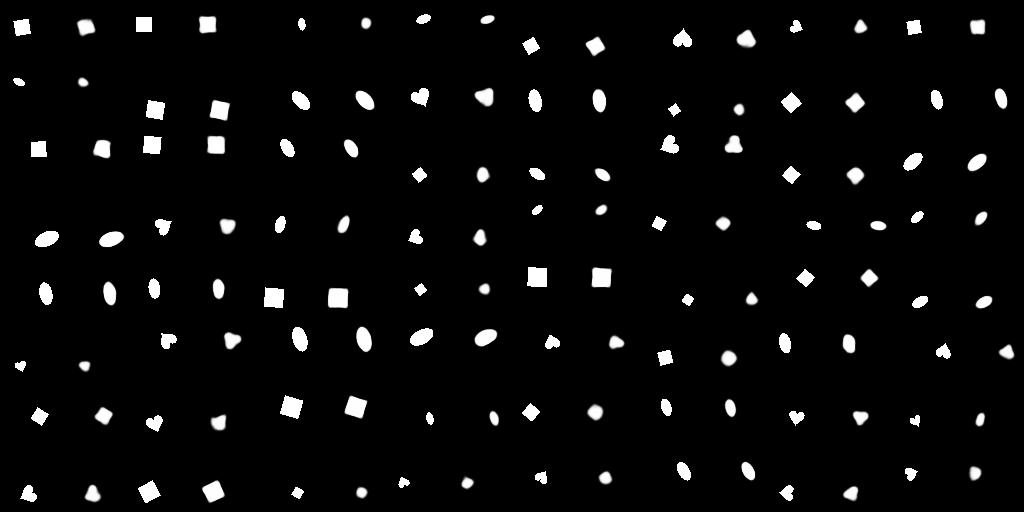}} \hspace{15em}%
    \subfloat[The reconstruction results of PRI-VAE on \textit{cars3d} data set.]{\includegraphics[width=.9\textwidth]{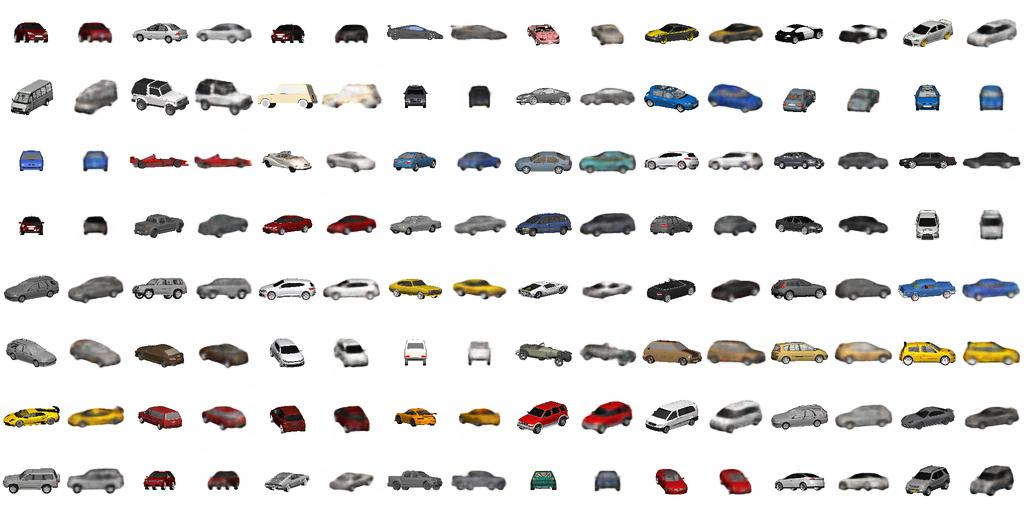}} 
    \caption{The reconstruction results of PRI-VAE model on (a) \textit{dsprites} and (b) \textit{cars3d} data sets. The odd columns represents the input data point and even columns are their reconstructions.
  PRI-VAE encourages more disentangled representations and reasonable reconstruction quality.} 
    \label{fig:privae_reconstruction}
\end{figure}

\end{document}